\documentclass[11pt]{article}

\usepackage[preprint]{acl}

\usepackage{times}
\usepackage{latexsym}

\usepackage[T1]{fontenc}

\usepackage[utf8]{inputenc}
\usepackage{multirow}

\usepackage{microtype}

\usepackage{inconsolata}
\usepackage{linguex}

\usepackage{graphicx}
 
\usepackage{booktabs}

\usepackage{adjustbox}
\usepackage{tcolorbox}
\usepackage{colortbl}
\tcbuselibrary{most}

\newtcolorbox{prompt}{
  enhanced,
  colback=gray!10,
  colframe=gray!50,
  arc=0mm,
  boxrule=0.5pt,
  left=5mm,
  right=5mm,
  top=2mm,
  bottom=2mm,
  fonttitle=\bfseries,
  coltitle=black,
  breakable,
  fontupper=\footnotesize\ttfamily,
      overlay={%
        \ifcase\tcbsegmentstate
        \or%
        \else%
        \fi%
    }
}
\usepackage{cleveref}

\title{Not all ANIMALs are equal: metaphorical framing through source domains and semantic frames}

\author{Yulia Otmakhova \qquad Matteo Guida \qquad Lea Frermann \\
  School of Computing and Information Systems,\\ The University of Melbourne \\
    \texttt{\href{mailto:y.otmakhova@unimelb.edu.au}{y.otmakhova@unimelb.edu.au}}, \\
    \texttt{guida@student.unimelb.edu.au, lea.frermann@unimelb.edu.au}
}

\begin{document}
\maketitle
\begin{abstract}

Metaphors are powerful framing devices, yet their source domains alone do not fully explain the specific associations they evoke. We argue that the interplay between source domains and semantic frames determines how metaphors shape understanding of complex issues, and present a computational framework that allows to derive salient discourse metaphors through their source domains and semantic frames. Applying this framework to climate change news, we uncover not only well-known source domains but also reveal nuanced frame-level associations that distinguish how the issue is portrayed. In analyzing immigration discourse across political ideologies, we demonstrate that liberals and conservatives systematically employ different semantic frames within the same source domains, with conservatives favoring frames emphasizing uncontrollability and liberals choosing neutral or more ``victimizing'' semantic frames. Our work bridges conceptual metaphor theory and linguistics, providing the first NLP approach for discovery of discourse metaphors and fine-grained analysis of differences in metaphorical framing.\footnote{Code, data and statistical scripts are available at  \url{https://github.com/julia-nixie/ConceptFrameMet}.}

\end{abstract}

\section{Introduction}

Metaphors help us understand and explain our world by transferring what we know about physical, tangible objects to more abstract, hard-to-define concepts and notions, or, as \citet{lakoff2008metaphors} succinctly put it, “understanding and experiencing one kind of thing in terms of another”. In particular, \citet{lakoff2008metaphors} showed that \textbf{target} (more abstract) concepts are understood in terms of more concrete, physical \textbf{source} domains, such as LOVE IS WAR or THEORIES ARE BUILDINGS. Following their seminal work, \textit{conceptual} metaphor theory (CMT) became a productive method to generalize from specific metaphorical expressions and arrive at a more abstract analysis of associations carried over from the source domain. In NLP, work on metaphorical understanding also focuses on mapping metaphors to their source domains 
\citep{shutova2010metaphor,mohler2016introducing,mendelsohn-budak-2025-people}.

However, the source domain itself does not fully explain which associations are carried over. 
Consider the following metaphors, both deriving from the source domain of WATER, commonly used in immigration discourse\footnote{Simplified examples from \citet{mendelsohn-budak-2025-people}'s dataset.}:

\ex. \label{ex:flood}
    Illegal aliens continue to \textit{flood} into our country, ruining our economy.

\ex. \label{ex:tide} Maybe a high \textit{tide} raises all boats?  \textit{Waves} of immigrants have always enriched us.

\begin{figure}
    \centering
    \includegraphics[width=0.96\linewidth]{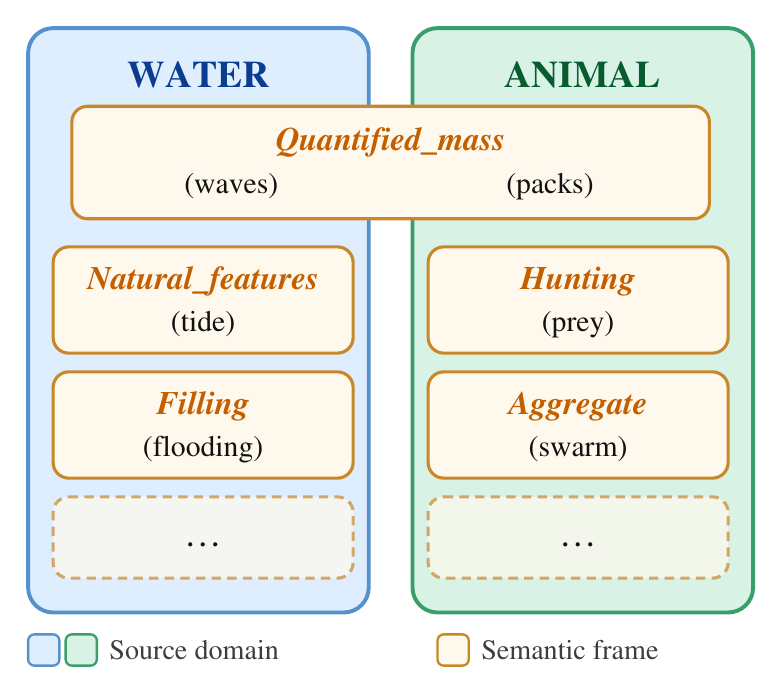}
    \caption{Interaction between semantic frames and domains}
    \label{fig:frames_domains}
\end{figure}
The first example compares immigration to \textit{filling} with water, which must be controlled and stopped; the second treats it as a \textit{mass of water} which, however, does not pose a risk of ``overflowing'' and can be considered a natural \textit{feature of landscape}, thus instilling a more positive, accepting attitude towards immigration.

Such associations have been explained by linguistic, constructionalist theories of metaphors \citep{Sullivan_2025, sullivan2013frames}, positing that the associations derived from the source domain are related to the \textit{semantic frame} of a lexical item used in the metaphor~\citep{fillmore2001frame}. In our example, \textit{flood} has a semantic frame of \textit{Filling} which  emphasizes the movement of water and the negative result of such movement. On the other hand, \textit{wave} and \textit{tide} have semantic frames of \textit{Quantified\_mass} and \textit{Natural\_features} with more neutral associations. Semantic frames are generic structures that can be applied to many source domains. For example, the \textit{Quantified\_mass} semantic frame can also be applied to the ANIMAL source domain (\textit{packs of immigrants}) or PRESSURE/PHYSICAL\_BURDEN source domain (\textit{loads of illegals}). Thus, it is the interplay of the source domain and semantic frame that uniquely defines the associations of the metaphor, where the source domain points to a cluster of associations, and the semantic frame allows to pick out specific ones (\Cref{fig:frames_domains}).

This nuanced understanding of associations is particularly important when metaphors are being used as {\it framing devices} to emphasize and highlight a particular aspect of an issue or a debate, i.e. to frame it \citep{entman1993framing}.\footnote{The term \textit{frame} refers to (at least) two distinct concepts -- semantic frames as encoded in FrameNet \citep{fillmore2001frame}, and media frames, i.e. consistent emphasis of particular aspects of an issue to evoke specific associations in reader's mind \citep{entman1993framing}; see \citet {otmakhova-etal-2024-media,sullivan2023three} for the discussion of their relation to each other and distinctions. To avoid confusion, we use terms \textit{semantic frames} and \textit{media frames}. 
} Such ``relatively stable metaphorical projections that function as key framing devices within a particular discourse over a certain period of time'' which are termed \textit{discourse metaphors} \cite[p.~134]{nerlich2012metaphors} are the focus of our study. Specifically, we show that differences in metaphor use between political ideologies --- which has not been fully explained in prior work~\citep{mendelsohn2024theory, el2001metaphors} --- can be effectively analyzed through prevalent choice of semantic frames that funnel particular associations from the source domain.

Our contributions are as follows:

\begin{itemize}
    \item We are the first to employ FrameNet-style semantic frames to pinpoint the differences in associations carried through source domains, and to analyse metaphorical framing at the discourse level rather than ``local'', individual metaphors.
    \item To support the above, we implement a framework that consists of two components: a pre-trained language model that detects metaphors and predicts their semantic frame and source domain, and a statistical module that uses log-likelihood ratio estimation \citep{rayson-garside-2000-comparing} to highlight salient source domains and semantic frames of metaphors within a particular discourse. 
    \item We apply our framework across two domains showing it can be used to \textit{discover} prevalent discourse metaphors and fine-grained associations within a particular topic such as climate change, and to \textit{highlight nuanced differences} in superficially similar metaphor usage, such as usage of ANIMAL, WATER and WAR metaphors within the immigration discourse by liberals and conservatives.
    \item We release a new corpus of climate change metaphors annotated with source domains, and re-open the discussion on the optimal taxomony of source domains, proposing semantic frames as a principled way to derive it.
\end{itemize}

\section{Related work}


\paragraph{Metaphor analysis in NLP}
Metaphors have attracted significant attention, with an extensive body of work focusing on annotating metaphors and creating corpora \citep{group2007mip,steen2010method,boisson-etal-2025-metaphorshare}, automatically detecting them in text \citep{wang2025ckemi,zhang-liu-2022-metaphor,uduehi-bunescu-2024-expectation,tian-etal-2024-theory,reimann-scheffler-2025-using}, understanding \citep{tong-etal-2024-metaphor,ye2025unveiling}, generation \citep{joseph-etal-2023-newsmet,chakrabarty-etal-2021-mermaid,veale-2016-round}, and interpretation, such as the ability to handle them in inference and question answering tasks \citep{liu-etal-2022-testing,sanchez-bayona-agerri-2025-metaphor} (see \citet{ge2023survey,rai2020survey} for an extensive survey).

Concept metaphor theory (CMT; \citet{lakoff2008metaphors}) has been widely adopted as a tool to improve detection, explanation, and generation of metaphors \citep{jones1992generating,stowe-etal-2021-exploring,ge2022explainable,mao-etal-2023-metapro,tian-etal-2025-imara,wachowiak-gromann-2023-gpt}. Some works use semantic frames as a proxy for conceptual mappings, in order to detect \citep{li-etal-2023-framebert}, or generate \citep{stowe-etal-2021-metaphor} metaphors. We are the first to leverage the interaction between the two frameworks. 

According to the constructionist view of metaphors, their conceptual understanding is bound by linguistic constructs \citep{sullivan2013frames,Sullivan_2025}. \citet{rosen-2018-computationally} use construction grammar cues such as argument structure to improve prediction of source domains of metaphors, \citet{jang-etal-2017-finding} collect syntactical patterns associated with a particular source domain for improved metaphor detection, while several studies encode semantic roles or domains ({\it agent}, {\it physical\_afflication}) 
\citep{gordon-etal-2015-corpus,dodge-etal-2015-metanet}, establishing schema that consist of multiple lexical instantiations. However, with the exception of a small descriptive study which looks at the differences in semantic frames used for the same CMT metaphor across two languages \citep{gamonal-2022-descriptive}, we are the first to use FrameNet-style semantic frames to contrast specific metaphor instantiations and pinpoint the differences in associations carried through them.

\paragraph{Metaphors as framing devices}
While metaphors are well-known framing devices \citep{landau2009evidence,boeynaems2017effects,semino2018integrated,brugman2019metaphorical}, the link between metaphors and framing is still under-explored in NLP. Several studies approximated framing effects of metaphors through other variables: \citet{prabhakaran2021metaphors} analyzed effects of metaphor usage on reader engagement, while \citet{baleato-rodriguez-etal-2023-paper} modeled metaphors to improve propaganda detection.

\begin{figure}[t]
   \centering
   \includegraphics[clip, trim=0.4cm 0 1cm 0, width=1\linewidth]{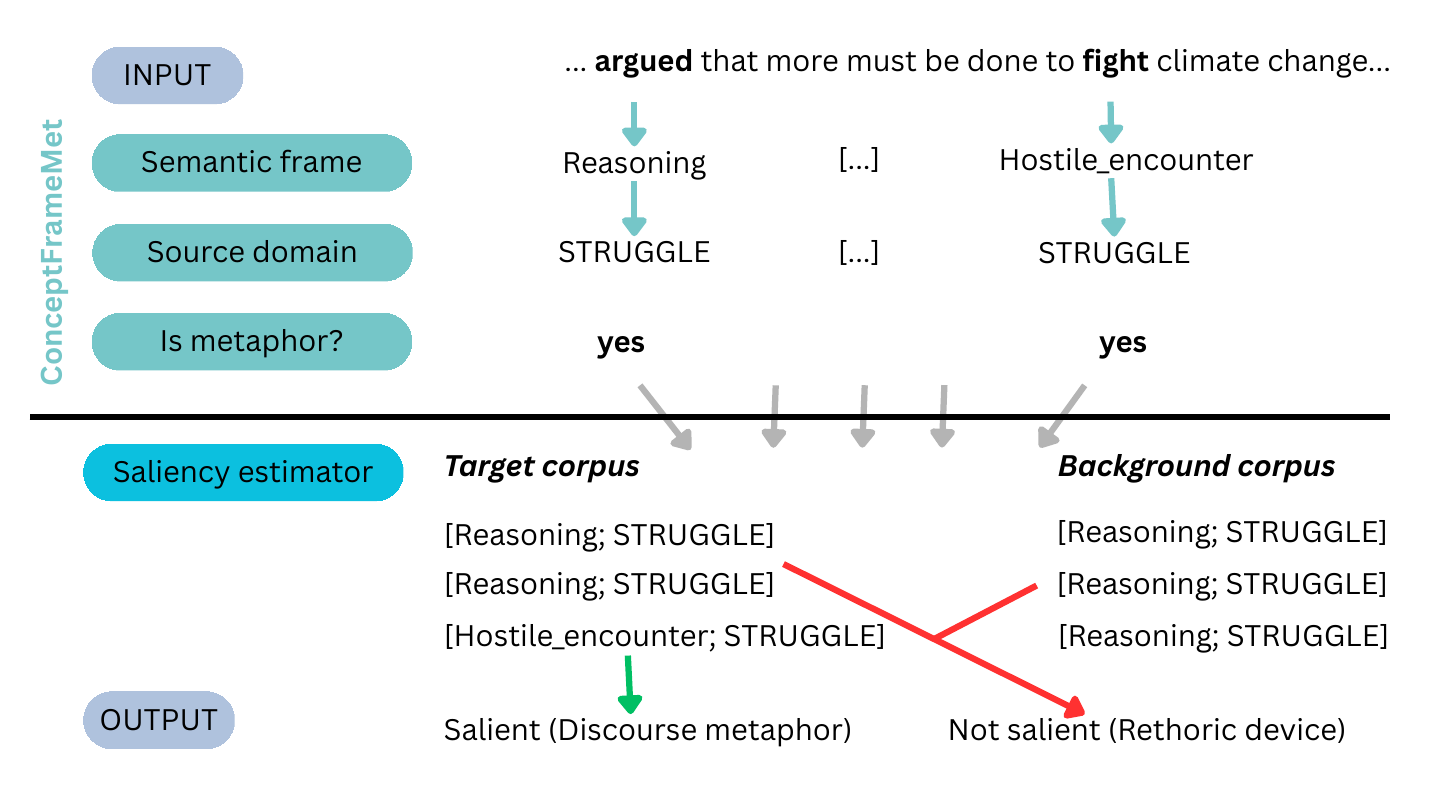}
   \caption{Framework overview with semantic frames and source domains for the input metaphors (\textbf{bold}).}
   \label{fig:model}
\end{figure}

A number of works focus on metaphors as framing devices by analysing their source domains
in a particular topic or discourse \citep{chen-etal-2022-framing,ahrens2022referential,sengupta-etal-2022-back,guan-zeng-2024-changes,li-etal-2024-evolving,meng2025large} or by statistically comparing source domains across political leanings~\citep{sengupta-etal-2024-analyzing,wang-2024-metaphorical,mendelsohn-budak-2025-people}. None of these studies, however, explain why metaphors from the same source domain are used across polar opposite ideologies~\citep{mendelsohn-budak-2025-people}, or considered finer associations within the source domain.
We bridge this gap by adopting semantic frames to extract particular associations from source domains, and saliency as a tool to discover prevalent and distinctive combinations of source domains and semantic frames.

\section{Methods}
\label{sec:methods}
We introduce our two-part framework: a model that combines metaphor detection with predicting their source domain and underlying semantic frames (\Cref{sec:prediction}), and the log-likelihood ratio method used to estimate the saliency of source domains and semantic frames in the corpus (\Cref{sec:saliency}). \Cref{fig:model} shows an overview of the framework. 

\subsection{Predicting metaphors, source domains and semantic frames}
\label{sec:prediction}

In this section we describe ConceptFrameMet -- a RoBERTa \citep{liu2019roberta} model for detecting metaphors and predicting their source domain and semantic frames (\Cref{fig:model}, top). The model relies on three components -- Semantic frame classifier (\Cref{sec:sem_frames}), Source domain classifier (\Cref{sec:source_domains}), and metaphor classifier (\Cref{sec:met_pred}).

\subsubsection{Semantic frame classifier}
\label{sec:sem_frames}

\textbf{Data.} We use FrameNet 1.7 \citep{ruppenhofer2016framenet} which contains sentences with semantic frames annotations over 797 classes, with the train, dev, and test splits of 19391/2272/6714 samples \citep{swayamdipta2017frame}.

\noindent\textbf{Models.} We fine-tune a RoBERTa-base model \citep{liu2019roberta} to output a probability distribution over 797 semantic frames for a target word in a sentence (details in \Cref{app:sem_frame_prediction}). We compare two input configurations: the target word masked out (\textit{immigrants MASK the cities}), and the target word separated from the context sentence (\textit{SEP immigrants flood the cities SEP flood SEP}). 
We also prompt two LLMs -- Gemini 2.5 and Claude Sonnet 4.0 -- to choose one of 797 classes (\Cref{app:sem_frame_prompt}). We include results reported on the same test set for the same task in two recent papers \citep{an-etal-2023-coarse,devasier-etal-2024-robust}, which, however, require substantial data-augmentation and are thus unsuitable for our large-scale topic-specific analyses.


\noindent\textbf{Results.} 
As shown in \Cref{tab:frame_prediction}, our best performing model (RoBERTa SEP) performs comparatively to the knowledge-heavy models in prior work, while LLMs significantly under-perform. The relatively low macro-F1 score is explained by a large number of minor classes which are semantically similar (i.e. \textit{Capability} vs \textit{Possibility}; \Cref{app:confusion_sem_frames} lists the most often confused semantic frames). However, their overall proportion is low and thus unlikely to significantly affect the downstream analysis. We use the SEP-style model in all further experiments.


\begin{table}[t]
    \centering
    \small
    \begin{tabular}{lccc}
        \toprule
        & \textbf{Acc} & \textbf{micro-F1} & \textbf{macro-F1}\\
        \midrule
        RoBERTa MASK & 0.806 & 0.806 & 0.053  \\
        RoBERTa SEP & \textbf{0.861} & \textbf{0.866} & \textbf{0.648} \\
        Gemini 2.5 & 0.508 & 0.508 & 0.430 \\
        Claude Sonnet 4.0 & 0.736 & 0.736 & 0.600 \\
        \midrule 
        \citet{an-etal-2023-coarse} & 0.899 & -- & -- \\
        \citet{devasier-etal-2024-robust} & -- & 0.887 & -- \\
        \bottomrule
    \end{tabular}
    \caption{Semantic frame prediction performance on FN1.7 test across three runs (sd < 0.05).
    }
    \label{tab:frame_prediction}
\end{table}

\subsubsection{Source domain classifier}
\label{sec:source_domains}

\textbf{Data.} We use the LCC Metaphor Dataset (large) \citep{mohler2016introducing}, the largest English dataset  annotated for source domains, and pre-process it as described in \ref{app:source_domain_prediction}. The resulting dataset has 99 source domains, and is randomly sampled with stratification into train, dev, and test of 11704/2509/2509 samples. 

\noindent\textbf{Models.} We fine-tune a RoBERTa-base model to predict one of 99 source domains (details in \ref{app:source_domain_prediction}). Similar to \Cref{sec:sem_frames}, we compare the masked input style with the inputs where the metaphor and the context are separated by SEP. Moreover, since some of the source domain classes are semantically close and easily confused \citep{shutova2010metaphor,mohler2016introducing}, we hypothesize that including the semantic frame of the metaphor will help the model to differentiate between domains. To test this, we pass the predicted probability distribution of semantic frames as a frozen feature vector to the model in two ways. In Frames\_CONCAT, we simply concatenate this feature vector with the RoBERTa-encoded input vector. In Frames\_ATTN, we maintain two copies of semantic frame vectors. The first is frozen, and the second is updated during training. During inference, we apply attention over the trainable matrix (with source domain embeddings as query) to highlight those semantic frames that are important for the source domain prediction, and add the frozen frame vectors as residuals. We also compare the fine-tuned models with zero-shot Gemini 2.5 and Claude Sonnet 4.0, where the prompt includes the list of 99 classes (\Cref{app:source_domain_prompt}). 

\noindent\textbf{Results.}  \Cref{tab:source_domain_results} confirms that adding semantic frame information boosts source domain prediction performance, with the attention-based model performing best overall. In particular, they achieve an improvement of 20 points of macro F1 on underrepresented classes (<10 instances in training data; see Appendix \Cref{fig:binned_results}). Most fine-tuned RoBERTa models again outperform zero-shot LLMs. We use the Frames\_ATTN model in all further analyses.

\begin{table}[t]
    \centering
    \small

    \begin{tabular}{lcccc}
        \toprule
        & \textbf{Acc} & \textbf{P} & \textbf{R} & \textbf{F1}\\
        \midrule
        RoBERTa MASK & 0.307 & 0.203 & 0.184 & 0.182 \\
        RoBERTa SEP & 0.833 & 0.745 & 0.742 & 0.740 \\
        Frames\_CONCAT & 0.837 & 0.759 & \textbf{0.758} & 0.754 \\
        Frames\_ATTN & \textbf{0.838} & \textbf{0.764} & 0.757 & \textbf{0.756}  \\
        Gemini 2.5 & 0.528 & 0.419 & 0.356 & 0.345 \\
        Claude Sonnet 4.0 & 0.528 & 0.517 & 0.452 & 0.445\\
        \bottomrule
        \end{tabular}
         \caption{Source domain prediction performance on the LCC dataset across three runs (sd < 0.05). All metrics are macro-averaged.}
    \label{tab:source_domain_results}

\end{table}




\subsubsection{Metaphor prediction}
\label{sec:met_pred}

\textbf{Data.} We use VUA-18, the largest available metaphor prediction dataset \citep{leong-etal-2018-report} to train and test our models, and evaluate generalizability through zero-shot testing on two smaller benchmarks -- MOH-X \citep{mohammad-etal-2016-metaphor} and TroFi \citep{birke-sarkar-2006-clustering} (see \Cref{app:metaphor_prediction} for data statistics).

\noindent\textbf{Models.} We integrate our semantic frame and source domain classifiers with a metaphor prediction model. For the latter, we choose MelBert \citep{choi2021melbert}, which is the strongest pre-trained model among ``knowledge-lean'' models that do not require augmenting data with additional features. 
MelBert is a RoBERTa model which  
captures the inconsistency between the contextual meaning of the metaphor and its literal meaning through contrasting embeddings of the word in context and in isolation. To improve on that, we posit that the encoding of a word in isolation still mixes representations of literal and metaphorical usage; for example, the embedding of the isolated word \textit{flooded} will likely reflect both the senses of \textit{the river flooded the city} and \textit{immigrants flooded the city}, i.e. still be somewhat close to a metaphorical context embedding. 
Thus in ConceptFrameMet we replace the word embedding with that of the predicted source domain (WATER), amplifying the difference between contexts in the metaphorical case (\textit{immigrants} and \textit{water}), and minimizing the gap in literal cases (\textit{river} and \textit{water}). To account for source domain prediction noise, instead of replacing a word with its predicted source domain, we blend their embeddings, using the confidence score of the source domain prediction ($\alpha$) as a weight: 
\begin{equation*}
\mathbf{e}_{\text{blended}} = \alpha \cdot \mathbf{e}_{\text{source\_domain}} + (1 - \alpha) \cdot \mathbf{e}_{\text{word}}.
\end{equation*}

We compare the resulting model with FrameBert \citep{li-etal-2023-framebert}, a modified version of MelBert, 
and Claude Sonnet 4.0 and Gemini 2.5 which are prompted to identify all metaphors (\Cref{app:met_pred_prompt}). To ensure fair comparison with pre-trained models, we only evaluate the results on target words.\footnote{For positive cases (metaphor present), we consider an LLM prediction to be correct if the target word belongs to a extracted span; for negative cases (no metaphor), we confirm that the target word does not occur in any of extracted spans.}

\begin{table}[t]
    \centering
    \small
    \begin{tabular}{lccc}
        \toprule
        & \multicolumn{1}{c}{\textbf{VUA-18}} & \multicolumn{1}{c}{\textbf{TroFi}} & \multicolumn{1}{c}{\textbf{MOH-X}} \\
        \midrule
        Random baseline & 0.222 & 0.466 & 0.486 \\
        \midrule
        MelBert & \textbf{0.782} & 0.631  & 0.806 \\
        FrameBert & 0.766 &0.620 & 0.780 \\
        Gemini 2.5 & 0.341 & 0.633 & \textbf{0.849} \\
        Claude Sonnet 4.0 & 0.330 & \textbf{0.651} &0.822 \\\midrule
        ConceptFrameMet (ours) & \underline{0.767} & \underline{0.634} & {0.814} \\
        \bottomrule
    \end{tabular}
    \caption{Binary F1 (metaphor class) performance on metaphor prediction for models fine-tuned on VUA-18 (train). We show results on VUA-18 (test) and two other test sets. LLM results are zero-shot for all three datasets. Bold=best; underline=second best.}
    \label{tab:metaphor_results}
\end{table}

\noindent\textbf{Results.} Results in \Cref{tab:metaphor_results,tab:full_metaphor_results} show that MelBert performs best in-domain (i.e., when tested on the test portion of the same data set) but fails to generalize to other data sets. LLMs perform best on less varied, verb-only test sets. ConcepFrameMet (our method) shows the best overall performance, only slightly underperforming the best method across all three test sets, while being substantially more resource efficient than the LLMs. The LLMs tend to over-predict metaphors, so they struggle on VUA-18 where the rate of metaphors is low (22\%). However, the results on VUA-18 are indicative of practical usage scenarios where one needs to distinguish metaphors from mostly literal words. This, together with low performance of LLMs on semantic frame and source domain prediction, motivates us to use our light-weight, integrated PLM model for metaphor detection and analysis.



\subsection{Detecting discourse metaphors}
\label{sec:saliency}

Since this study focuses on metaphors as a media framing device, we distinguish between \textit{discourse metaphors} --- which carry stable, prevalent associations used to frame some issue \citep{scheufele2010spreading} --- and metaphors that are merely figures of speech. Unfortunately, they cannot be differentiated based on their raw counts in text since some metaphors are used frequently across different contexts. Rather, a metaphor has to be salient in a given discourse \citep{zinken2008discourse}, i.e. occur in it more frequently than in other discourses (see lower half of \Cref{fig:model}). Unlike \citet{mason2004cormet} who compare raw metaphor frequencies between corpora, we operationalize \textit{saliency} as the log-likelihood ratio \citep{rayson-garside-2000-comparing}, which is a common technique to compare distributions of items (such as words or, in our case, semantic frames and source domains) between two corpora. In particular, given two corpora $C_1$ and $C_2$, the log-likelihood ratio identifies candidate metaphors that are significantly over-represented in one of the corpora, i.e., reject the null hypothesis $H_0$ that they are represented equally in both corpora ($C_1$ and $C_2$).
\[
-2\ln \lambda = -2[\ell(\theta_0) - \ell(\hat{\theta})] = 2\sum O_i \ln\left(\frac{O_i}{E_i}\right)
\]
\noindent where $\lambda$ = likelihood ratio, $\ell(\theta_0)$ = log-likelihood under $H_0$, 
$\ell(\hat{\theta})$ = maximum log-likelihood (from ($C_1$, $C_2$)), $O_i$ = observed frequency, 
$E_i$ = expected frequency under $H_0$.

Source domains and semantic frames with log-likelihood values that reject $H_0$ with $p>0.05$ are identified and their relative frequency is compared to decide which of the corpora ($C_1$ or $C_2$) they are more strongly associated with.

\section{Metaphorical framing through source domains and semantic frames}

We now apply our framework explained in \Cref{sec:methods} to (1)~\textit{discover} discourse metaphors in texts about a particular topic by comparing metaphors in a topic-specific vs. a general back ground corpus (\Cref{sec:climate_change}); and \textit{contrast} discourse metaphors in texts associated with polar opposite political leanings within a single topic (\Cref{sec:immi}). 

\subsection{Metaphorical framing in climate change news}
\label{sec:climate_change}

Existing methods for metaphorical framing analysis relies on manual discovery of prevalent source domains in a particular discourse, while NLP approaches require such knowledge of source domains a priori. Here, we apply our framework to automatically identify salient source domains from a large collection of news on climate change, and analyze them through salient semantic frames.

\begin{table}[t]
\centering
\small
\begin{tabular}{lcccc}
\toprule
& \textbf{Acc} & \textbf{P} & \textbf{R} & \textbf{F1} \\
\midrule
Metaphor prediction & 0.94 & 0.94 & 1 & 0.97 \\
Source domain prediction & 0.88 & 0.62 & 0.66 & 0.64 \\
\bottomrule
\end{tabular}
         \caption{Performance of ConceptFrameMet on the climate dataset. Scores are binary for metaphor detection, and macro-averaged for source domain prediction.}
    \label{tab:climate_results}
\end{table}
\subsubsection{Data and methods}

We collect a dataset of 47K paragraphs about climate change from a corpus of New York Times articles published between 1986 and 2020 \citep{fast2017long,mendelsohn2020framework} and an equivalent generic corpus randomly sampled from the same source (details in \Cref{app:term_distrubution}).
Next, we use ConceptFrameMet (\Cref{sec:prediction}) to detect metaphors, their semantic frames and source domains, in both the climate and generic corpus. To ensure that the identified metaphors refer to the topic of interest, we retain only sentences that contain terms ``climate'' or ``warming''. Finally, we calculate log-likelihood ratios for source domains, and then for semantic frames within each source domain, as described in \Cref{sec:saliency}.

\begin{table*}[t]
    \centering
    \small
    \setlength{\tabcolsep}{4pt}

    \label{tab:placeholder_label}
    \begin{tabular}{c | c c | c c}
        \hline
        & \multicolumn{2}{c}{\textbf{Climate corpus}} & \multicolumn{2}{c}{\textbf{Generic corpus}}\\
        & Semantic frame & Metaphors & Semantic frame & Metaphors \\
        \midrule
        \multirow{3}{*}{STRUGGLE}  & \textit{Hostile\_encounter}   & fight, confront & \textit{Difficulty} & challenges, hard  \\
        & \textit{Topic}  & address   & \textit{Resolve\_problem}  & settle \\
        & \textit{Relation} & at odds & \textit{Cause\_to\_end} & ended \\
        \hline
        
        \multirow{3}{*}{WAR}  & \textit{Hostile\_encounter}  &  battle, waged war  & \textit{Change\_of\_leadership} & rebel, revolution \\
        & \textit{Boundary}  & on the front lines   & \textit{Invading}  & invasion, intervene \\
        & \textit{Judgment\_communication} & assailed, crusaded & \textit{Aiming} & target \\
        \hline

        \multirow{3}{*}{OBJECT HANDLING}  & \textit{Attempt\_suasion}  &  push (for), press  & \textit{Getting} & gets \\
        & \textit{Taking\_sides}  & take (a stand), handle   & \textit{Intercepting}  & snapping up, caught \\
        & \textit{Intentionally\_act} & take (action, steps) & \textit{Entity} & things, stuff \\
        \hline

        \multirow{3}{*}{HUMAN BODY}  & \textit{Confronting\_problem}  &  face, confront  & \textit{Social\_connection} & contacts, close \\
        & \textit{Taking\_sides}  & stance, embrace   & \textit{Body\_parts}  & eye for, at the wrist \\
        & \textit{Part\_orientational} & at the heart of  & \textit{First\_rank} & core  \\
        \hline

    \end{tabular}
     \caption{Examples of salient semantic frames (with SOURCE DOMAINS and Metaphors) in the climate change corpus (left) and the generic corpus (right).}
     \label{tab:examples}
\end{table*}

\subsubsection{Evaluation}
The process above resulted in a corpus of 6,859 sentences that contain metaphors with source domains and semantic frames salient in the climate change corpus ($p<0.05$). To evaluate the performance of ConceptFrameMet on that corpus, we randomly select a sample of 416 sentences that covers all combinations of salient source domains and semantic frames within them. We presented crowd workers on Prolific with sentences containing highlighted metaphors and asked them to identify up to three source domains from a list of 5 options, comprising the top two domains predicted by ConceptFrameMet and three random distractors plus options OTHER DOMAIN and NO METAPHOR (see \Cref{app:annotation_task} for more details).
We collect four annotators per sample and achieve a reliable averaged pair-wise annotator agreement of 68\% 
, and use majority voting (with adjudication by one author this paper with a degree in linguistics) to determine the final source domain label.
We finally compute model performance against human labels, showing that ConceptFrameMet generalizes well to our new domain (\Cref{tab:climate_results}).\footnote{Note that the metaphor prediction recall and F1 are inflated since all samples were predicted as containing metaphors by the model; in this we follow the process and limitations of the metaphor corpus creation suggested by \citet{mohler2016introducing}.}

\begin{figure}
    \centering
    \includegraphics[clip, trim=0.4cm 0 0.5cm 0, width=1\linewidth]{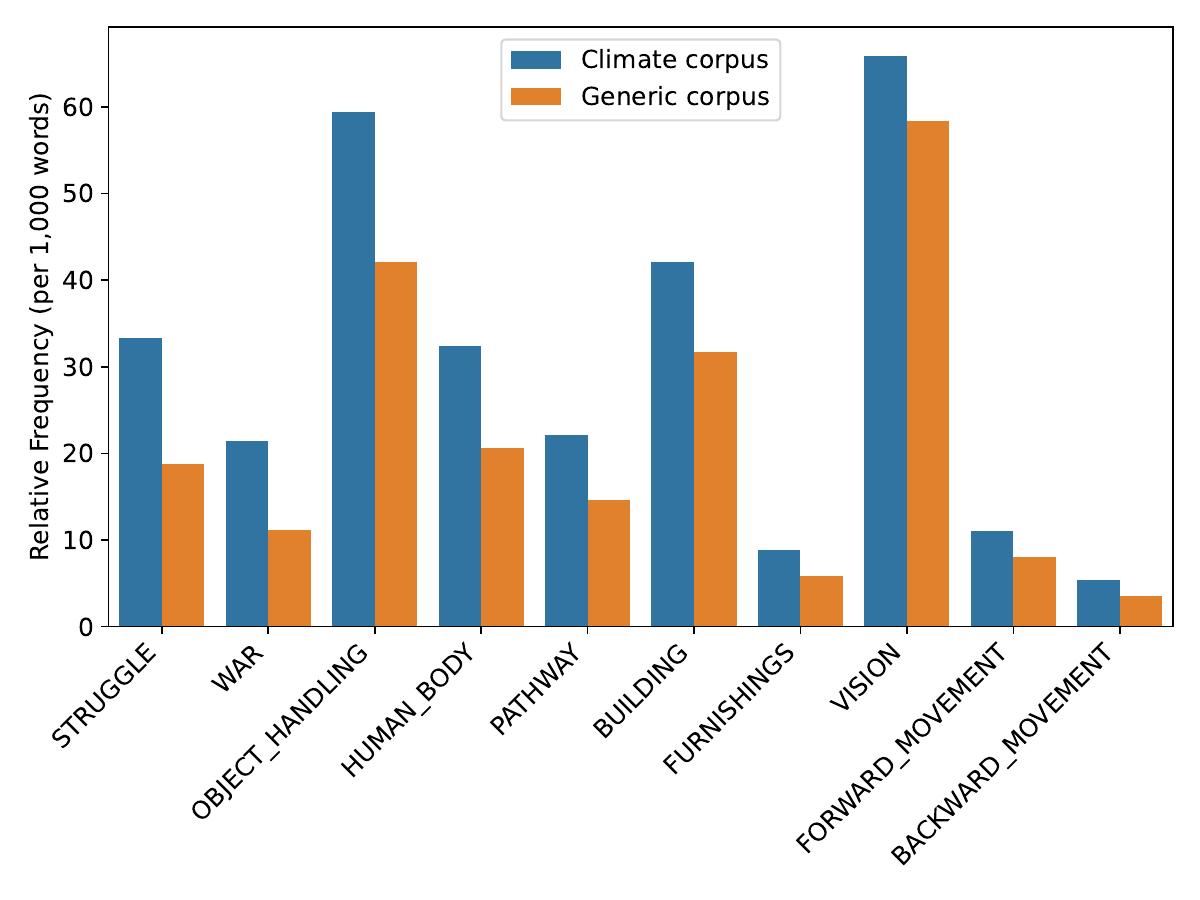}
    \caption{Relative frequency of source domains in climate and generic corpora. The source domains are sorted left to right in order of diminishing saliency}
    \label{fig:salient_domains}
\end{figure}

\subsubsection{Results}

\paragraph{Source domains}
\Cref{fig:salient_domains} presents the 10 most salient source domains in the climate corpus, ordered by log-likelihood ratio, from left to right, while the y-axis indicates frequency, noting that saliency highlights \textit{differences in frequency} of source domains between corpora rather than the \textit{absolute frequency} of a source domain within a corpus. Thus, a frequent source domain (such as VISION) can be ranked lower than a less frequent domain (STRUGGLE), and the most frequent source domain in the climate corpus (MOVEMENT) is not among the most salient ones. Importantly, the discovered source domains are well-aligned with the ones reported in theoretical critical discourse analysis studies. We present examples of metaphors from each of the top 10 source domains, together with the theoretical studies, in \Cref{app:top_source_domains}.

\paragraph{Semantic frames}
Even within a salient source domain, some metaphors can be used as a figure of speech rather than a discourse (framing) metaphor. For example, the source domain STRUGGLE (which compares mental hardships to physical ones such as fighting) includes metaphorical use of \textit{argue} (\textit{The ministry argued that...}) which are used in the climate corpus with similar frequency as in general texts, as well as climate-specific STRUGGLE metaphors such as \textit{fight} ({\it more must be done to fight climate change}). 

When comparing a topic-focused corpus against a randomized, generic corpus, log-likelihood ratio estimation allows to achieve two things: first, metaphors that are common figures of speech, used equally frequently irrespective of the text's topic, which will have a statistically insignificant differences. Second, it points towards prominent {\it framings} of the topic (as repeatedly used metaphors), as well as absent (but possible) framings (which metaphors are more rare than in general usage). 

\Cref{tab:examples} shows examples of semantic frames for top source domains associated with climate change which are salient in climate discourse (left) or are avoided (right). In some source domains, this difference is particularly revealing -- for example, climate metaphors in the OBJECT HANDLING source domain tend to use semantic frames denoting intentional actions such as \textit{Attempt\_suasion} (push for) or rather than more ``passive'' semantic frames such as  \textit{Intercepting} (caught). Interestingly, while the metaphors on both sides of the WAR source domain express fierce activity, climate change actions are not framed as a revolution (\textit{Change\_of\_leadership}) or intervention (\textit{Invading}). 

In sum, semantic frames as mental templates allow to pick out specific parts of a source domain that correspond to the ways we relate to and act upon the climate crisis. However, they cannot represent the metaphorical frame on their own -- as \Cref{tab:examples} shows, the same semantic frame (for example \textit{Taking\_sides}) can be used across different source domains, imbued with different physical imagery. Thus, discovering prevalent metaphorical frames requires both components.

\subsection{Conservative vs liberal framing in immigration discourse on Twitter/X}
\label{sec:immi}

Previous research has shown that, perhaps surprisingly, the same dehumanizing source domains are used to frame immigration by both conservatives and liberals, albeit with different frequency and intensity~\cite{mendelsohn-budak-2025-people}. 
Here, we study the interplay of source domains and semantic frames to shed light on the more nuanced differences of metaphor use across political camps.

\subsubsection{Data and methods}

We use a corpus of 400K US tweets on immigration with automatically predicted metaphoricity scores and scores of political affiliation (liberal vs conservative; \citet{mendelsohn-budak-2025-people}. We select three well-studied domains (WATER, ANIMAL and WAR) for analysis.

We extract and filter out candidate tweets as explained in \Cref{app:immi_source_domains}, and use ConceptFrameMet to annotate them for metaphors, their semantic frames and source domains (the resulting statistics in \Cref{tab:immi_source_domains}). We apply log-likelihood ratio to semantic frames within the ANIMAL,WATER and WAR source domains to highlight those which are more representative of liberal vs conservative discourse. To do so, we split each subset of tweets with metaphors of a particular source domain into two parts according to their predicted ideology, and use them as $C_1$ and $C_2$, respectively. 

\subsubsection{Results}

\begin{figure}
    \centering
    \includegraphics[width=1\linewidth]{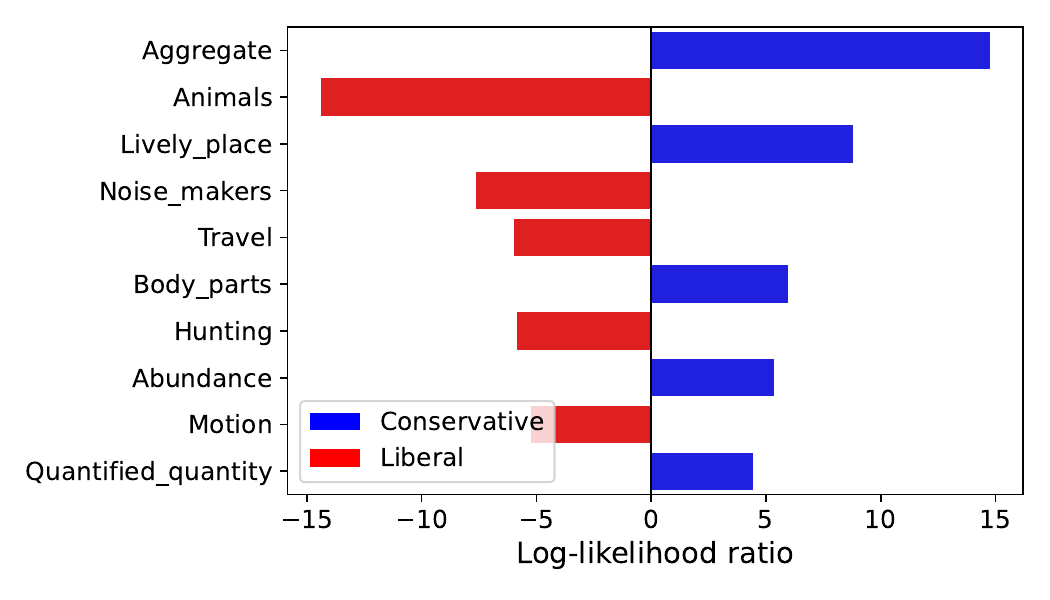}
    \caption{The five semantic frames with highest saliency in the liberal (red) and conservative (blue) data, in the ANIMAL domain.}
    \label{fig:lib_cons}
\end{figure}

While many semantic frames are used by both political leanings (e.g. the top most used semantic frame in the ANIMAL source domain is \textit{Animals} for both sides, which is a reference to an animal or the word ``animal''), their \textit{saliency} differs. \Cref{fig:lib_cons} shows the top five most salient ANIMAL semantic frames for both leanings. The semantic frame \textit{Animals} is salient in liberal tweets, where it is used to criticize dehumanization (``Families shouldn't be ripped apart and treated like \textbf{animals}''). Other prevalent ANIMAL metaphors in liberal tweets are based on semantic frames of \textit{Hunting} (``ICE agents are \textbf{predators}, undocumented immigrants are \textbf{prey}''), 
or \textit{Moral\_evaluation} (``How is that not \textbf{predatory}?''). Conversely, the ANIMAL metaphors in conservative tweets are predominately using semantic frames like \textit{Aggregate} (``Anytime you disrupt a rats \textbf{nest} [...] they scatter''; ``the \textbf{swarm} of Muslim migrants''), \textit{Body parts} (``ILLEGALS taking [...] every benefit they can lay their \textbf{paws} on''), or \textit{Abundance} (``the illegal immigrants will run \textbf{rampant}'').

Overall, we observe that the semantic frames used by conservatives pick out more specific, ``colorful'', but also more negative parts of the ANIMAL source domain, while liberals often use semantic frames that show lack of agency and power, with immigrants portrayed as victims or passive (i.e. immigrants are \textit{caged}, \textit{hunted}, \textit{treated like animals}).  

Within the WATER source domain, the conservative side features only a few salient semantic frames like \textit{Filling} (``\textbf{Flooding} America with illegals''; ``those cities won’t take \textbf{overflow} Illegal Immigrants'') and \textit{Abounding with} (``It’s now a \textbf{cesspool} of crime'', ``CA is \textbf{saturated} with illegals''), which depict immigration as ``too much'' that needs to be controlled. On the other hand, the semantic frames on the liberal side have more neutral associations: \textit{Quantified mass} (``the \textbf{trickle} of immigrants''),  \textit{Natural feature} (``Many different immigrant \textbf{streams}''), or, again, frame immigrants as powerless victims through such semantic frames as \textbf{Killing} and \textbf{Death} (``the federales tried to \textbf{drown} families'') or \textit{Removing} (``this [policy] is in part to \textbf{flush out} immigrant parents''). Similarly, the WAR metaphors on the conservative side demonstrate semantic paucity (despite their frequent usage), with only use 2 semantic frames of \textit{Invading} and \textit{Arriving} (``invaders'', ``invasion'' etc). On the other hand, liberals have a broader variety of semantic frames highlighting their persecution by authorities such as \textit{Hostile\_encounter} (``waging war''), \textit{Attack} (``raid'', ``assault'', ``onslaught''), \textit{Revenge} (``crackdown'', ``retaliate'') etc). 

\section{Discussion}

We highlight several open-ended questions regarding metaphor analysis that fall out of our work.

\paragraph{Can semantic frames provide a more principled way of choosing source domains?}
To date, there exists no principled way of constructing the ontology of conceptual mappings between source and target domains. The Master
Metaphor List \citep{lakoff1991master} has been criticized for conflating concepts across different taxonomy levels, resulting in categories that are neither consistent in granularity nor mutually exclusive \citep{lonneker2008hamburg, shutova2010metaphor}. We also observe this in the LCC corpus \citep{mohler2016introducing}, and estimate which source domains are likely to be problematic using our annotated corpus (see \Cref{app:annotation_task}). We calculate normalized point-wise mutual information (NPMI) of pairs of labels that are chosen for the same metaphor by a single annotator, or by all annotators. To account for noise, we weight the resulting score by the frequency of a particular label pair (\Cref{app:confusion}). This highlights such source domains as BATTLE vs WAR, BARRIER vs PATHWAY, and LIGHT vs VISION as being the hardest to distinguish (top 10 confusions in \Cref{tab:npmi_within,tab:npmi_between}).

The distribution of semantic frames that support these competing source domains, however, reveals two distinct patterns. Some of the source domains such as BATTLE are completely subsumed by another domain (WAR): all four semantic frames of BATTLE are also part of the WAR source domain (65 semantic frames). Thus, BATTLE is a child of the WAR source domain, and can be considered redundant. On the other hand, while some source domains are related in meaning (BARRIER vs PATHWAY), the bulk of semantic frames constituting each of them is distinct: only 30\% of PATHWAY semantic frames overlap with BARRIER, and 45\% in the reverse direction. Here, the composition of source domains could help to choose a more appropriate source domain where multiple mappings are possible. For example, the metaphor \textit{roadblocks (to net zero)} is closer to such semantic frames as \textit{Hindering}, \textit{Prohibiting},  \textit{Thwarting} that belong to BARRIER domain, rather than to \textit{Means}, \textit{Manner}, \textit{Trajectory} that characterize PATHWAYS domain. Thus, semantic frames can help to establish the relationship between the source domains and provide a more systematic way of distinguishing between them.   

\paragraph{Can the semantic analysis of source domains be reproduced by LLMs end-to-end?}

 While the results above demonstrate sub-par LLM performance on the individual tasks in our pipeline (metaphor detection, source domain prediction, semantic frame identification), this does not rule out the possibility that LLMs are able to perform the full task end-to-end and identify the most salient metaphors within the same instruction workflow. To test this, we conducted an experiment on two subsets of the immigration tweet corpus, namely those with source domain WATER or ANIMAL. Thus, we simplify the task, only asking the model to identify salient semantic frames.

We used Claude Sonnet 4.0 with instructions that combined all steps of the pipeline (detect metaphors, identify semantic frames, decide which semantic frames are salient for the conservative vs liberal side). We provided the list of semantic frames, and supplied the sets of liberal and conservatives tweets for a particular domain, clearly marking them as such (prompt in \Cref{app:end_to_end_prompt}). 

We found the results lacking in multiple respects. Some semantic frames (e.g. ``dehumanization'' for the ANIMAL domain, \Cref{app:end_to_end_animal}) and the examples to support them were hallucinated. The LLM also attributed the same lexical item (for example, ``flood'') to multiple semantic frames, or the same semantic frames (``Catastrophe'') to both liberals and conservatives (\Cref{app:end_to_end_water1}), unable to compare and weigh the semantic frames in a principled way. Most importantly, we suspect that LLM predictions are strongly affected by their pre-training knowledge, and not primarily driven by the input corpus. To test this, we presented the LLM with tweets that were unlikely to include WATER metaphors but prompted it to identify WATER metaphors (\Cref{app:end_to_end_water2}). The model still generated semantic frames, and they were nearly identical to those generated when given context, though the examples generated by the model either lacked WATER metaphors or were not found in the provided context (\Cref{app:end_to_end_water1}). This suggests that the model has apriori ``knowledge'' about metaphors and retrofits the examples to it, either picking them from the texts (if they are available) or hallucinating them (when they are not). For large-scale analyses it is unclear how to check for hallucination and/or outputs that are not grounded in the discourse of interest, how to rank the source domains and semantic frames in a reproducible way, and how to objectively judge the results considering that the model also adds its interpretation which, while superficially plausible, is detrimental to a principled analysis. In contrast, the advantages of our approach are interpretability, grounding in existing resources, and the ability to assess the statistical significance of the findings. Our work also provides an important reference for future work to benchmark and improve LLMs to make their analysis more grounded and less interpretative.

\section{Conclusions}

Inspired by a constructionist view of metaphor theory, which posits that metaphors are bounded by their linguistic expression, we proposed a framework, ConceptFrameMet, that models metaphors through both their source domains and semantic frames and show how salient semantic frames within source domains allow to capture nuanced differences in framing evoked by the metaphors. We release a pre-trained model with predicts metaphors, their source domains and semantic frames, as well as a statistical model for estimating the saliency of source domains and semantic frames within a particular topic or discourse. We demonstrate the potential of our framework by applying it to two different tasks -- \textit{discovering} source domains and semantic frames used in framing climate change news, and \textit{highlighting nuanced differences} in semantic frames used by liberals and conservatives within the same metaphorical source domain. 

\section{Limitations}

Our study has several limitations. First, we use a model trained on a set of source domains used in the LCC corpus \citep{mohler2016introducing}, which might not be comprehensive enough to cover new topics that are very different from the ones occurring in it. Thus, a larger -- or even open -- set of source domains is needed, and a method which is sensitive enough to differentiate between a large number of source domains. Relatedly, the question of distinguishing between related or semantically similar source domains \citep{shutova2010metaphor} is still unanswered, though in \Cref{sec:source_domains} we show an example of how semantic frames can serve as a tool to make such distinctions in a more principled way. Finally, in this study we only look at individual semantic frames within a particular domain, and do not consider how they interplay and link with each other in the same document to create more powerful extended metaphors. Thus, we hope to connect our FrameNet-frames based analysis -- which operates at lexical level -- with image-schematic approaches \citep{wachowiak-etal-2022-drum} or the MetaNet \citep{dodge-etal-2015-metanet} framework, which encode the relationships between the components of the source domain, to enable more sophisticated analysis of semantic frames that support complex, extended discourse metaphors.

\section*{Acknowledgments}
This work was supported by the Australian Research Council Discovery Early Career Research Award (Grant No.\ DE230100761) and University of Melbourne's Early Career Researcher Grant (Grant No.\ 2025ECRG137).

\bibliography{custom}

@inproceedings{otmakhova-etal-2024-media,
    title = "Media Framing: A typology and Survey of Computational Approaches Across Disciplines",
    author = "Otmakhova, Yulia  and
      Khanehzar, Shima  and
      Frermann, Lea",
    editor = "Ku, Lun-Wei  and
      Martins, Andre  and
      Srikumar, Vivek",
    booktitle = "Proceedings of the 62nd Annual Meeting of the Association for Computational Linguistics (Volume 1: Long Papers)",
    month = aug,
    year = "2024",
    address = "Bangkok, Thailand",
    publisher = "Association for Computational Linguistics",
    url = "https://aclanthology.org/2024.acl-long.822/",
    doi = "10.18653/v1/2024.acl-long.822",
    pages = "15407--15428"
}

@article{entman1993framing,
  title={Framing: Towards clarification of a fractured paradigm},
  author={Entman, Robert M},
  journal={McQuail's reader in mass communication theory},
  volume={390},
  pages={397},
  year={1993}
}

@article{sullivan2023three,
  title={Three levels of framing},
  author={Sullivan, Karen},
  journal={Wiley Interdisciplinary Reviews: Cognitive Science},
  volume={14},
  number={5},
  pages={e1651},
  year={2023},
  publisher={Wiley Online Library}
}

@phdthesis{mendelsohn2024theory,
  title={Theory-grounded Computational Analysis of Political Framing in Online Media},
  author={Mendelsohn, Julia},
  year={2024},
  school={University of Michigan},
  type={{PhD} dissertation}
}

@inproceedings{mendelsohn-budak-2025-people,
    title = "When People are Floods: Analyzing Dehumanizing Metaphors in Immigration Discourse with Large Language Models",
    author = "Mendelsohn, Julia  and
      Budak, Ceren",
    editor = "Che, Wanxiang  and
      Nabende, Joyce  and
      Shutova, Ekaterina  and
      Pilehvar, Mohammad Taher",
    booktitle = "Proceedings of the 63rd Annual Meeting of the Association for Computational Linguistics (Volume 1: Long Papers)",
    month = jul,
    year = "2025",
    address = "Vienna, Austria",
    publisher = "Association for Computational Linguistics",
    url = "https://aclanthology.org/2025.acl-long.398/",
    doi = "10.18653/v1/2025.acl-long.398",
    pages = "8079--8103",
    ISBN = "979-8-89176-251-0"
}

@book{lakoff2008metaphors,
  title={Metaphors we live by},
  author={Lakoff, George and Johnson, Mark},
  year={2008},
  publisher={University of Chicago press}
}

@inproceedings{shutova2010metaphor,
  title={Metaphor Corpus Annotated for Source-Target Domain Mappings.},
  author={Shutova, Ekaterina and Teufel, Simone},
  booktitle={LREC},
  volume={2},
  number={2},
  pages={2--2},
  year={2010}
}

@inproceedings{mohler2016introducing,
  title={Introducing the {LCC} metaphor datasets},
  author={Mohler, Michael and Brunson, Mary and Rink, Bryan and Tomlinson, Marc},
  booktitle={Proceedings of the Tenth International Conference on Language Resources and Evaluation (LREC'16)},
  pages={4221--4227},
  year={2016}
}

@book{sullivan2013frames,
  title={Frames and constructions in metaphoric language},
  author={Sullivan, Karen},
  year={2013},
  publisher={John Benjamins Publishing Company}
}

@inbook{Sullivan_2025, place={Cambridge}, series={Cambridge Handbooks in Language and Linguistics}, title={Metaphors and Constructions}, booktitle={The Cambridge Handbook of Construction Grammar}, publisher={Cambridge University Press}, author={Sullivan, Karen}, editor={Fried, Mirjam and Nikiforidou, KikiEditors}, year={2025}, pages={129–146}, collection={Cambridge Handbooks in Language and Linguistics}}

@inproceedings{fillmore2001frame,
  title={Frame semantics for text understanding},
  author={Fillmore, Charles J and Baker, Collin F},
  booktitle={Proceedings of WordNet and Other Lexical Resources Workshop, NAACL},
  volume={6},
  pages={59--64},
  year={2001}
}

@inproceedings{sengupta-etal-2022-back,
    title = "Back to the Roots: Predicting the Source Domain of Metaphors using Contrastive Learning",
    author = "Sengupta, Meghdut  and
      Alshomary, Milad  and
      Wachsmuth, Henning",
    editor = "Ghosh, Debanjan  and
      Beigman Klebanov, Beata  and
      Muresan, Smaranda  and
      Feldman, Anna  and
      Poria, Soujanya  and
      Chakrabarty, Tuhin",
    booktitle = "Proceedings of the 3rd Workshop on Figurative Language Processing (FLP)",
    month = dec,
    year = "2022",
    address = "Abu Dhabi, United Arab Emirates (Hybrid)",
    publisher = "Association for Computational Linguistics",
    url = "https://aclanthology.org/2022.flp-1.19/",
    doi = "10.18653/v1/2022.flp-1.19",
    pages = "137--142"
}

@inproceedings{jang-etal-2017-finding,
    title = "Finding Structure in Figurative Language: Metaphor Detection with Topic-based Frames",
    author = "Jang, Hyeju  and
      Maki, Keith  and
      Hovy, Eduard  and
      Ros{\'e}, Carolyn",
    editor = "Jokinen, Kristiina  and
      Stede, Manfred  and
      DeVault, David  and
      Louis, Annie",
    booktitle = "Proceedings of the 18th Annual {SIG}dial Meeting on Discourse and Dialogue",
    month = aug,
    year = "2017",
    address = {Saarbr{\"u}cken, Germany},
    publisher = "Association for Computational Linguistics",
    url = "https://aclanthology.org/W17-5538/",
    doi = "10.18653/v1/W17-5538",
    pages = "320--330"
}

@article{liu2019roberta,
  title={{RoBERTa}: A robustly optimized {BERT} pretraining approach},
  author={Liu, Yinhan and Ott, Myle and Goyal, Naman and Du, Jingfei and Joshi, Mandar and Chen, Danqi and Levy, Omer and Lewis, Mike and Zettlemoyer, Luke and Stoyanov, Veselin},
  journal={arXiv preprint arXiv:1907.11692},
  year={2019}
}

@incollection{scheufele2010spreading,
  title={Of spreading activation, applicability, and schemas: Conceptual distinctions and their operational implications for measuring frames and framing effects},
  author={Scheufele, Bertram T and Scheufele, Dietram A},
  booktitle={{Doing News Framing Analysis}},
  pages={126--150},
  year={2010},
  publisher={Routledge}
}

@inproceedings{choi2021melbert,
  title={{MelBERT}: Metaphor Detection via Contextualized Late Interaction using Metaphorical Identification Theories},
  author={Choi, Minjin and Lee, Sunkyung and Choi, Eunseong and Park, Heesoo and Lee, Junhyuk and Lee, Dongwon and Lee, Jongwuk},
  booktitle={Proceedings of the 2021 Conference of the North American Chapter of the Association for Computational Linguistics: Human Language Technologies},
  pages={1763--1773},
  year={2021}
}

@article{group2007mip,
  title={{MIP}: A method for identifying metaphorically used words in discourse},
  author={Group, Pragglejaz},
  journal={Metaphor and symbol},
  volume={22},
  number={1},
  pages={1--39},
  year={2007},
  publisher={Taylor \& Francis}
}

@article{wang2025ckemi,
  title={{CKEMI}: Concept knowledge enhanced metaphor identification framework},
  author={Wang, Dian and Li, Yang and Wang, Suge and Chen, Xin and Liao, Jian and Li, Deyu and Li, Xiaoli},
  journal={Information Processing \& Management},
  volume={62},
  number={1},
  pages={103946},
  year={2025},
  publisher={Elsevier}
}

@inproceedings{leong-etal-2018-report,
    title = "A Report on the 2018 {VUA} Metaphor Detection Shared Task",
    author = "Leong, Chee Wee (Ben)  and
      Beigman Klebanov, Beata  and
      Shutova, Ekaterina",
    editor = "Beigman Klebanov, Beata  and
      Shutova, Ekaterina  and
      Lichtenstein, Patricia  and
      Muresan, Smaranda  and
      Wee, Chee",
    booktitle = "Proceedings of the Workshop on Figurative Language Processing",
    month = jun,
    year = "2018",
    address = "New Orleans, Louisiana",
    publisher = "Association for Computational Linguistics",
    url = "https://aclanthology.org/W18-0907/",
    doi = "10.18653/v1/W18-0907",
    pages = "56--66"
}

@inproceedings{birke-sarkar-2006-clustering,
    title = "A Clustering Approach for Nearly Unsupervised Recognition of Nonliteral Language",
    author = "Birke, Julia  and
      Sarkar, Anoop",
    editor = "McCarthy, Diana  and
      Wintner, Shuly",
    booktitle = "11th Conference of the {E}uropean Chapter of the Association for Computational Linguistics",
    month = apr,
    year = "2006",
    address = "Trento, Italy",
    publisher = "Association for Computational Linguistics",
    url = "https://aclanthology.org/E06-1042/",
    pages = "329--336"
}

@inproceedings{mohammad-etal-2016-metaphor,
    title = "Metaphor as a Medium for Emotion: An Empirical Study",
    author = "Mohammad, Saif  and
      Shutova, Ekaterina  and
      Turney, Peter",
    editor = "Gardent, Claire  and
      Bernardi, Raffaella  and
      Titov, Ivan",
    booktitle = "Proceedings of the Fifth Joint Conference on Lexical and Computational Semantics",
    month = aug,
    year = "2016",
    address = "Berlin, Germany",
    publisher = "Association for Computational Linguistics",
    url = "https://aclanthology.org/S16-2003/",
    doi = "10.18653/v1/S16-2003",
    pages = "23--33"
}

@inproceedings{li-etal-2023-framebert,
    title = "{F}rame{BERT}: Conceptual Metaphor Detection with Frame Embedding Learning",
    author = "Li, Yucheng  and
      Wang, Shun  and
      Lin, Chenghua  and
      Guerin, Frank  and
      Barrault, Loic",
    editor = "Vlachos, Andreas  and
      Augenstein, Isabelle",
    booktitle = "Proceedings of the 17th Conference of the European Chapter of the Association for Computational Linguistics",
    month = may,
    year = "2023",
    address = "Dubrovnik, Croatia",
    publisher = "Association for Computational Linguistics",
    url = "https://aclanthology.org/2023.eacl-main.114/",
    doi = "10.18653/v1/2023.eacl-main.114",
    pages = "1558--1563"
}

@inproceedings{rayson-garside-2000-comparing,
    title = "Comparing Corpora using Frequency Profiling",
    author = "Rayson, Paul  and
      Garside, Roger",
    booktitle = "The Workshop on Comparing Corpora",
    month = oct,
    year = "2000",
    address = "Hong Kong, China",
    publisher = "Association for Computational Linguistics",
    url = "https://aclanthology.org/W00-0901/",
    doi = "10.3115/1117729.1117730",
    pages = "1--6"
}

@inproceedings{fast2017long,
  title={Long-term trends in the public perception of artificial intelligence},
  author={Fast, Ethan and Horvitz, Eric},
  booktitle={Proceedings of the AAAI conference on artificial intelligence},
  volume={31},
  number={1},
  year={2017}
}

@article{mendelsohn2020framework,
  title={A framework for the computational linguistic analysis of dehumanization},
  author={Mendelsohn, Julia and Tsvetkov, Yulia and Jurafsky, Dan},
  journal={Frontiers in artificial intelligence},
  volume={3},
  pages={55},
  year={2020},
  publisher={Frontiers Media SA}
}

@article{webersinke2021climatebert,
  title={{ClimateBERT}: A pretrained language model for climate-related text},
  author={Webersinke, Nicolas and Kraus, Mathias and Bingler, Julia Anna and Leippold, Markus},
  journal={arXiv preprint arXiv:2110.12010},
  year={2021}
}

@techreport{bingler2023cheaptalk,
    title={How Cheap Talk in Climate Disclosures Relates to Climate Initiatives, Corporate Emissions, and Reputation Risk},
    author={Bingler, Julia and Kraus, Mathias and Leippold, Markus and Webersinke, Nicolas},
    type={Working paper},
    institution={Available at SSRN 3998435},
    year={2023}
}

@article{zinken2008discourse,
  title={Discourse metaphors},
  author={Zinken, J{\"o}rg and Hellsten, Iina and Nerlich, Brigitte},
  journal={Body, language and mind},
  volume={2},
  pages={363--385},
  year={2008},
  publisher={John Benjamins Amsterdam}
}

@article{nerlich2012metaphors,
  title={Metaphors we die by? {G}eoengineering, metaphors, and the argument from catastrophe},
  author={Nerlich, Brigitte and Jaspal, Rusi},
  journal={Metaphor and symbol},
  volume={27},
  number={2},
  pages={131--147},
  year={2012},
  publisher={Taylor \& Francis}
}

@article{el2001metaphors,
  title={Metaphors we discriminate by: Naturalized themes in {Austrian} newspaper articles about asylum seekers},
  author={El Refaie, Elisabeth},
  journal={Journal of Sociolinguistics},
  volume={5},
  number={3},
  pages={352--371},
  year={2001},
  publisher={Wiley Online Library}
}

@article{steen2010method,
  title={A method for linguistic metaphor identification},
  author={Steen, Gerard J and Dorst, Aletta G and Krennmayr, Tina and Kaal, Anna A and Herrmann, J Berenike},
  year={2010},
  publisher={John Benjamins Publishing Company}
}

@inproceedings{sanchez-bayona-agerri-2025-metaphor,
    title = "Metaphor and Large Language Models: When Surface Features Matter More than Deep Understanding",
    author = "Sanchez-Bayona, Elisa  and
      Agerri, Rodrigo",
    editor = "Che, Wanxiang  and
      Nabende, Joyce  and
      Shutova, Ekaterina  and
      Pilehvar, Mohammad Taher",
    booktitle = "Findings of the Association for Computational Linguistics: ACL 2025",
    month = jul,
    year = "2025",
    address = "Vienna, Austria",
    publisher = "Association for Computational Linguistics",
    url = "https://aclanthology.org/2025.findings-acl.898/",
    doi = "10.18653/v1/2025.findings-acl.898",
    pages = "17462--17477",
    ISBN = "979-8-89176-256-5"
}

@inproceedings{reimann-scheffler-2025-using,
    title = "Using Large Language Models to Perform {MIPVU}-Inspired Automatic Metaphor Detection",
    author = "Reimann, Sebastian  and
      Scheffler, Tatjana",
    editor = "Rambelli, Giulia  and
      Ilievski, Filip  and
      Bolognesi, Marianna  and
      Sommerauer, Pia",
    booktitle = "Proceedings of the 2nd Workshop on Analogical Abstraction in Cognition, Perception, and Language (Analogy-Angle II)",
    month = aug,
    year = "2025",
    address = "Vienna, Austria",
    publisher = "Association for Computational Linguistics",
    url = "https://aclanthology.org/2025.analogyangle-1.2/",
    doi = "10.18653/v1/2025.analogyangle-1.2",
    pages = "10--21",
    ISBN = "979-8-89176-274-9"
}

@inproceedings{zhang-liu-2022-metaphor,
    title = "Metaphor Detection via Linguistics Enhanced {S}iamese Network",
    author = "Zhang, Shenglong  and
      Liu, Ying",
    editor = "Calzolari, Nicoletta  and
      Huang, Chu-Ren  and
      Kim, Hansaem  and
      Pustejovsky, James  and
      Wanner, Leo  and
      Choi, Key-Sun  and
      Ryu, Pum-Mo  and
      Chen, Hsin-Hsi  and
      Donatelli, Lucia  and
      Ji, Heng  and
      Kurohashi, Sadao  and
      Paggio, Patrizia  and
      Xue, Nianwen  and
      Kim, Seokhwan  and
      Hahm, Younggyun  and
      He, Zhong  and
      Lee, Tony Kyungil  and
      Santus, Enrico  and
      Bond, Francis  and
      Na, Seung-Hoon",
    booktitle = "Proceedings of the 29th International Conference on Computational Linguistics",
    month = oct,
    year = "2022",
    address = "Gyeongju, Republic of Korea",
    publisher = "International Committee on Computational Linguistics",
    url = "https://aclanthology.org/2022.coling-1.364/",
    pages = "4149--4159"
}

@inproceedings{uduehi-bunescu-2024-expectation,
    title = "An Expectation-Realization Model for Metaphor Detection",
    author = "Uduehi, Oseremen  and
      Bunescu, Razvan",
    editor = "Ghosh, Debanjan  and
      Muresan, Smaranda  and
      Feldman, Anna  and
      Chakrabarty, Tuhin  and
      Liu, Emmy",
    booktitle = "Proceedings of the 4th Workshop on Figurative Language Processing (FigLang 2024)",
    month = jun,
    year = "2024",
    address = "Mexico City, Mexico (Hybrid)",
    publisher = "Association for Computational Linguistics",
    url = "https://aclanthology.org/2024.figlang-1.11/",
    doi = "10.18653/v1/2024.figlang-1.11",
    pages = "79--84"
}

@article{ge2023survey,
  title={A survey on computational metaphor processing techniques: From identification, interpretation, generation to application},
  author={Ge, Mengshi and Mao, Rui and Cambria, Erik},
  journal={Artificial Intelligence Review},
  volume={56},
  number={Suppl 2},
  pages={1829--1895},
  year={2023},
  publisher={Springer}
}

@article{rai2020survey,
  title={A survey on computational metaphor processing},
  author={Rai, Sunny and Chakraverty, Shampa},
  journal={ACM Computing Surveys (CSUR)},
  volume={53},
  number={2},
  pages={1--37},
  year={2020},
  publisher={ACM New York, NY, USA}
}

@inproceedings{liu-etal-2022-testing,
    title = "Testing the Ability of Language Models to Interpret Figurative Language",
    author = "Liu, Emmy  and
      Cui, Chenxuan  and
      Zheng, Kenneth  and
      Neubig, Graham",
    editor = "Carpuat, Marine  and
      de Marneffe, Marie-Catherine  and
      Meza Ruiz, Ivan Vladimir",
    booktitle = "Proceedings of the 2022 Conference of the North American Chapter of the Association for Computational Linguistics: Human Language Technologies",
    month = jul,
    year = "2022",
    address = "Seattle, United States",
    publisher = "Association for Computational Linguistics",
    url = "https://aclanthology.org/2022.naacl-main.330/",
    doi = "10.18653/v1/2022.naacl-main.330",
    pages = "4437--4452"
}

@inproceedings{boisson-etal-2025-metaphorshare,
    title = "{METAPHORSHARE}: A Dynamic Collaborative Repository of Open Metaphor Datasets",
    author = "Boisson, Joanne  and
      Mehmood, Arif  and
      Camacho-Collados, Jose",
    editor = "Dziri, Nouha  and
      Ren, Sean (Xiang)  and
      Diao, Shizhe",
    booktitle = "Proceedings of the 2025 Conference of the Nations of the Americas Chapter of the Association for Computational Linguistics: Human Language Technologies (System Demonstrations)",
    month = apr,
    year = "2025",
    address = "Albuquerque, New Mexico",
    publisher = "Association for Computational Linguistics",
    url = "https://aclanthology.org/2025.naacl-demo.41/",
    doi = "10.18653/v1/2025.naacl-demo.41",
    pages = "509--521",
    ISBN = "979-8-89176-191-9"
}

@inproceedings{stowe-etal-2021-metaphor,
    title = "Metaphor Generation with Conceptual Mappings",
    author = "Stowe, Kevin  and
      Chakrabarty, Tuhin  and
      Peng, Nanyun  and
      Muresan, Smaranda  and
      Gurevych, Iryna",
    editor = "Zong, Chengqing  and
      Xia, Fei  and
      Li, Wenjie  and
      Navigli, Roberto",
    booktitle = "Proceedings of the 59th Annual Meeting of the Association for Computational Linguistics and the 11th International Joint Conference on Natural Language Processing (Volume 1: Long Papers)",
    month = aug,
    year = "2021",
    address = "Online",
    publisher = "Association for Computational Linguistics",
    url = "https://aclanthology.org/2021.acl-long.524/",
    doi = "10.18653/v1/2021.acl-long.524",
    pages = "6724--6736"
}

@inproceedings{veale-2016-round,
    title = "Round Up The Usual Suspects: Knowledge-Based Metaphor Generation",
    author = "Veale, Tony",
    editor = "Beigman Klebanov, Beata  and
      Shutova, Ekaterina  and
      Lichtenstein, Patricia",
    booktitle = "Proceedings of the Fourth Workshop on Metaphor in {NLP}",
    month = jun,
    year = "2016",
    address = "San Diego, California",
    publisher = "Association for Computational Linguistics",
    url = "https://aclanthology.org/W16-1105/",
    doi = "10.18653/v1/W16-1105",
    pages = "34--41"
}

@inproceedings{joseph-etal-2023-newsmet,
    title = "{N}ews{M}et : A ``do it all'' Dataset of Contemporary Metaphors in News Headlines",
    author = "Joseph, Rohan  and
      Liu, Timothy  and
      Ng, Aik Beng  and
      See, Simon  and
      Rai, Sunny",
    editor = "Rogers, Anna  and
      Boyd-Graber, Jordan  and
      Okazaki, Naoaki",
    booktitle = "Findings of the Association for Computational Linguistics: ACL 2023",
    month = jul,
    year = "2023",
    address = "Toronto, Canada",
    publisher = "Association for Computational Linguistics",
    url = "https://aclanthology.org/2023.findings-acl.641/",
    doi = "10.18653/v1/2023.findings-acl.641",
    pages = "10090--10104"
}

@inproceedings{tian-etal-2024-theory,
    title = "A Theory Guided Scaffolding Instruction Framework for {LLM}-Enabled Metaphor Reasoning",
    author = "Tian, Yuan  and
      Xu, Nan  and
      Mao, Wenji",
    editor = "Duh, Kevin  and
      Gomez, Helena  and
      Bethard, Steven",
    booktitle = "Proceedings of the 2024 Conference of the North American Chapter of the Association for Computational Linguistics: Human Language Technologies (Volume 1: Long Papers)",
    month = jun,
    year = "2024",
    address = "Mexico City, Mexico",
    publisher = "Association for Computational Linguistics",
    url = "https://aclanthology.org/2024.naacl-long.428/",
    doi = "10.18653/v1/2024.naacl-long.428",
    pages = "7738--7755"
}

@inproceedings{chakrabarty-etal-2021-mermaid,
    title = "{MERMAID}: Metaphor Generation with Symbolism and Discriminative Decoding",
    author = "Chakrabarty, Tuhin  and
      Zhang, Xurui  and
      Muresan, Smaranda  and
      Peng, Nanyun",
    editor = "Toutanova, Kristina  and
      Rumshisky, Anna  and
      Zettlemoyer, Luke  and
      Hakkani-Tur, Dilek  and
      Beltagy, Iz  and
      Bethard, Steven  and
      Cotterell, Ryan  and
      Chakraborty, Tanmoy  and
      Zhou, Yichao",
    booktitle = "Proceedings of the 2021 Conference of the North American Chapter of the Association for Computational Linguistics: Human Language Technologies",
    month = jun,
    year = "2021",
    address = "Online",
    publisher = "Association for Computational Linguistics",
    url = "https://aclanthology.org/2021.naacl-main.336/",
    doi = "10.18653/v1/2021.naacl-main.336",
    pages = "4250--4261"
}

@inproceedings{tian-etal-2025-imara,
    title = "{I}ma{RA}: An Imaginative Frame Augmented Method for Low-Resource Multimodal Metaphor Detection and Explanation",
    author = "Tian, Yuan  and
      Wang, Minzheng  and
      Xu, Nan  and
      Mao, Wenji",
    editor = "Chiruzzo, Luis  and
      Ritter, Alan  and
      Wang, Lu",
    booktitle = "Findings of the Association for Computational Linguistics: NAACL 2025",
    month = apr,
    year = "2025",
    address = "Albuquerque, New Mexico",
    publisher = "Association for Computational Linguistics",
    url = "https://aclanthology.org/2025.findings-naacl.220/",
    doi = "10.18653/v1/2025.findings-naacl.220",
    pages = "3953--3967",
    ISBN = "979-8-89176-195-7"
}

@inproceedings{tong-etal-2024-metaphor,
    title = "Metaphor Understanding Challenge Dataset for {LLM}s",
    author = "Tong, Xiaoyu  and
      Choenni, Rochelle  and
      Lewis, Martha  and
      Shutova, Ekaterina",
    editor = "Ku, Lun-Wei  and
      Martins, Andre  and
      Srikumar, Vivek",
    booktitle = "Proceedings of the 62nd Annual Meeting of the Association for Computational Linguistics (Volume 1: Long Papers)",
    month = aug,
    year = "2024",
    address = "Bangkok, Thailand",
    publisher = "Association for Computational Linguistics",
    url = "https://aclanthology.org/2024.acl-long.193/",
    doi = "10.18653/v1/2024.acl-long.193",
    pages = "3517--3536"
}

@article{ye2025unveiling,
  title={Unveiling {LLMs}' Metaphorical Understanding: Exploring Conceptual Irrelevance, Context Leveraging and Syntactic Influence},
  author={Ye, Fengying and Wang, Shanshan and Chao, Lidia S and Wong, Derek F},
  journal={arXiv preprint arXiv:2510.04120},
  year={2025}
}

@inproceedings{jones1992generating,
  title={Generating a specific class of metaphors},
  author={Jones, Mark},
  booktitle={30th Annual Meeting of the Association for Computational Linguistics},
  pages={321--323},
  year={1992}
}

@inproceedings{mao-etal-2023-metapro,
    title = "{M}eta{P}ro Online: A Computational Metaphor Processing Online System",
    author = "Mao, Rui  and
      Li, Xiao  and
      He, Kai  and
      Ge, Mengshi  and
      Cambria, Erik",
    editor = "Bollegala, Danushka  and
      Huang, Ruihong  and
      Ritter, Alan",
    booktitle = "Proceedings of the 61st Annual Meeting of the Association for Computational Linguistics (Volume 3: System Demonstrations)",
    month = jul,
    year = "2023",
    address = "Toronto, Canada",
    publisher = "Association for Computational Linguistics",
    url = "https://aclanthology.org/2023.acl-demo.12/",
    doi = "10.18653/v1/2023.acl-demo.12",
    pages = "127--135"
}

@inproceedings{ge2022explainable,
  title={Explainable metaphor identification inspired by {Conceptual Metaphor Theory}},
  author={Ge, Mengshi and Mao, Rui and Cambria, Erik},
  booktitle={Proceedings of the AAAI conference on artificial intelligence},
  volume={36},
  number={10},
  pages={10681--10689},
  year={2022}
}

@inproceedings{stowe-etal-2021-exploring,
    title = "Exploring Metaphoric Paraphrase Generation",
    author = "Stowe, Kevin  and
      Beck, Nils  and
      Gurevych, Iryna",
    editor = "Bisazza, Arianna  and
      Abend, Omri",
    booktitle = "Proceedings of the 25th Conference on Computational Natural Language Learning",
    month = nov,
    year = "2021",
    address = "Online",
    publisher = "Association for Computational Linguistics",
    url = "https://aclanthology.org/2021.conll-1.26/",
    doi = "10.18653/v1/2021.conll-1.26",
    pages = "323--336"
}

@inproceedings{rosen-2018-computationally,
    title = "Computationally Constructed Concepts: A Machine Learning Approach to Metaphor Interpretation Using Usage-Based Construction Grammatical Cues",
    author = "Rosen, Zachary",
    editor = "Beigman Klebanov, Beata  and
      Shutova, Ekaterina  and
      Lichtenstein, Patricia  and
      Muresan, Smaranda  and
      Wee, Chee",
    booktitle = "Proceedings of the Workshop on Figurative Language Processing",
    month = jun,
    year = "2018",
    address = "New Orleans, Louisiana",
    publisher = "Association for Computational Linguistics",
    url = "https://aclanthology.org/W18-0912/",
    doi = "10.18653/v1/W18-0912",
    pages = "102--109"
}

@inproceedings{dodge-etal-2015-metanet,
    title = "{M}eta{N}et: Deep semantic automatic metaphor analysis",
    author = "Dodge, Ellen  and
      Hong, Jisup  and
      Stickles, Elise",
    editor = "Shutova, Ekaterina  and
      Beigman Klebanov, Beata  and
      Lichtenstein, Patricia",
    booktitle = "Proceedings of the Third Workshop on Metaphor in {NLP}",
    month = jun,
    year = "2015",
    address = "Denver, Colorado",
    publisher = "Association for Computational Linguistics",
    url = "https://aclanthology.org/W15-1405/",
    doi = "10.3115/v1/W15-1405",
    pages = "40--49"
}

@inproceedings{gordon-etal-2015-corpus,
    title = "A Corpus of Rich Metaphor Annotation",
    author = "Gordon, Jonathan  and
      Hobbs, Jerry  and
      May, Jonathan  and
      Mohler, Michael  and
      Morbini, Fabrizio  and
      Rink, Bryan  and
      Tomlinson, Marc  and
      Wertheim, Suzanne",
    editor = "Shutova, Ekaterina  and
      Beigman Klebanov, Beata  and
      Lichtenstein, Patricia",
    booktitle = "Proceedings of the Third Workshop on Metaphor in {NLP}",
    month = jun,
    year = "2015",
    address = "Denver, Colorado",
    publisher = "Association for Computational Linguistics",
    url = "https://aclanthology.org/W15-1407/",
    doi = "10.3115/v1/W15-1407",
    pages = "56--66"
}

@inproceedings{gamonal-2022-descriptive,
    title = "A Descriptive Study of Metaphors and Frames in the Multilingual Shared Annotation Task",
    author = "Gamonal, Maucha",
    editor = "Baker, Collin F.",
    booktitle = "Proceedings of the Workshop on Dimensions of Meaning: Distributional and Curated Semantics (DistCurate 2022)",
    month = jul,
    year = "2022",
    address = "Seattle, Washington",
    publisher = "Association for Computational Linguistics",
    url = "https://aclanthology.org/2022.distcurate-1.1/",
    doi = "10.18653/v1/2022.distcurate-1.1",
    pages = "1--7"
}

@inproceedings{wang-2024-metaphorical,
    title = "Metaphorical Framing of Refugees, Asylum Seekers and Immigrants in {UK}'s Left and Right-Wing Media",
    author = "Wang, Yunxiao",
    editor = "Bizzoni, Yuri  and
      Degaetano-Ortlieb, Stefania  and
      Kazantseva, Anna  and
      Szpakowicz, Stan",
    booktitle = "Proceedings of the 8th Joint SIGHUM Workshop on Computational Linguistics for Cultural Heritage, Social Sciences, Humanities and Literature (LaTeCH-CLfL 2024)",
    month = mar,
    year = "2024",
    address = "St. Julians, Malta",
    publisher = "Association for Computational Linguistics",
    url = "https://aclanthology.org/2024.latechclfl-1.3/",
    pages = "18--27"
}

@article{landau2009evidence,
  title={Evidence that self-relevant motives and metaphoric framing interact to influence political and social attitudes},
  author={Landau, Mark J and Sullivan, Daniel and Greenberg, Jeff},
  journal={Psychological science},
  volume={20},
  number={11},
  pages={1421--1427},
  year={2009},
  publisher={SAGE Publications Sage CA: Los Angeles, CA}
}

@article{brugman2019metaphorical,
  title={Metaphorical framing in political discourse through words vs. concepts: A meta-analysis},
  author={Brugman, Britta C and Burgers, Christian and Vis, Barbara},
  journal={Language and Cognition},
  volume={11},
  number={1},
  pages={41--65},
  year={2019},
  publisher={Cambridge University Press}
}

@article{boeynaems2017effects,
  title={The effects of metaphorical framing on political persuasion: A systematic literature review},
  author={Boeynaems, Amber and Burgers, Christian and Konijn, Elly A and Steen, Gerard J},
  journal={Metaphor and Symbol},
  volume={32},
  number={2},
  pages={118--134},
  year={2017},
  publisher={Taylor \& Francis}
}

@article{semino2018integrated,
  title={An integrated approach to metaphor and framing in cognition, discourse, and practice, with an application to metaphors for cancer},
  author={Semino, Elena and Demj{\'e}n, Zs{\'o}fia and Demmen, Jane},
  journal={Applied linguistics},
  volume={39},
  number={5},
  pages={625--645},
  year={2018},
  publisher={Oxford University Press}
}

@inproceedings{prabhakaran2021metaphors,
  title={How metaphors impact political discourse: A large-scale topic-agnostic study using neural metaphor detection},
  author={Prabhakaran, Vinodkumar and Rei, Marek and Shutova, Ekaterina},
  booktitle={Proceedings of the International AAAI Conference on Web and Social Media},
  volume={15},
  pages={503--512},
  year={2021}
}

@inproceedings{baleato-rodriguez-etal-2023-paper,
    title = "Paper Bullets: Modeling Propaganda with the Help of Metaphor",
    author = "Baleato Rodr{\'i}guez, Daniel  and
      Dankers, Verna  and
      Nakov, Preslav  and
      Shutova, Ekaterina",
    editor = "Vlachos, Andreas  and
      Augenstein, Isabelle",
    booktitle = "Findings of the Association for Computational Linguistics: EACL 2023",
    month = may,
    year = "2023",
    address = "Dubrovnik, Croatia",
    publisher = "Association for Computational Linguistics",
    url = "https://aclanthology.org/2023.findings-eacl.35/",
    doi = "10.18653/v1/2023.findings-eacl.35",
    pages = "472--489"
}

@article{lakoff1991master,
  title={Master metaphor list (Technical report)},
  author={Lakoff, George and Espenson, Jane and Schwartz, Alan},
  journal={Cognitive Linguistics Group University of California, Berkeley},
  year={1991}
}

@inproceedings{sengupta-etal-2024-analyzing,
    title = "Analyzing the Use of Metaphors in News Editorials for Political Framing",
    author = "Sengupta, Meghdut  and
      El Baff, Roxanne  and
      Alshomary, Milad  and
      Wachsmuth, Henning",
    editor = "Duh, Kevin  and
      Gomez, Helena  and
      Bethard, Steven",
    booktitle = "Proceedings of the 2024 Conference of the North American Chapter of the Association for Computational Linguistics: Human Language Technologies (Volume 1: Long Papers)",
    month = jun,
    year = "2024",
    address = "Mexico City, Mexico",
    publisher = "Association for Computational Linguistics",
    url = "https://aclanthology.org/2024.naacl-long.199/",
    doi = "10.18653/v1/2024.naacl-long.199",
    pages = "3621--3631"
}

@inproceedings{guan-zeng-2024-changes,
    title = "Changes in the Sentiments and Metaphors in {COVID}-19 News Discourse (2019-2024)",
    author = "Guan, Yolanda  and
      Zeng, Winnie Huiheng",
    editor = "Oco, Nathaniel  and
      Dita, Shirley N.  and
      Borlongan, Ariane Macalinga  and
      Kim, Jong-Bok",
    booktitle = "Proceedings of the 38th Pacific Asia Conference on Language, Information and Computation",
    month = dec,
    year = "2024",
    address = "Tokyo, Japan",
    publisher = "Tokyo University of Foreign Studies",
    url = "https://aclanthology.org/2024.paclic-1.78/",
    pages = "810--819"
}

@inproceedings{chen-etal-2022-framing,
    title = "Framing Legitimacy in {CSR}: A Corpus of {C}hinese and {A}merican Petroleum Company {CSR} Reports and Preliminary Analysis",
    author = "Chen, Jieyu  and
      Ahrens, Kathleen  and
      Huang, Chu-Ren",
    editor = "Wan, Mingyu  and
      Huang, Chu-Ren",
    booktitle = "Proceedings of the First Computing Social Responsibility Workshop within the 13th Language Resources and Evaluation Conference",
    month = jun,
    year = "2022",
    address = "Marseille, France",
    publisher = "European Language Resources Association",
    url = "https://aclanthology.org/2022.csrnlp-1.4/",
    pages = "24--34"
}

@inproceedings{li-etal-2024-evolving,
    title = "The Evolving Use of {WAR} Metaphors in Businesswomen-focused Media Discourse",
    author = "Li, Yanlin  and
      Chen, Jing  and
      Ahrens, Kathleen  and
      Huang, Chu-Ren",
    editor = "Oco, Nathaniel  and
      Dita, Shirley N.  and
      Borlongan, Ariane Macalinga  and
      Kim, Jong-Bok",
    booktitle = "Proceedings of the 38th Pacific Asia Conference on Language, Information and Computation",
    month = dec,
    year = "2024",
    address = "Tokyo, Japan",
    publisher = "Tokyo University of Foreign Studies",
    url = "https://aclanthology.org/2024.paclic-1.135/",
    pages = "1377--1386"
}

@article{meng2025large,
  title={Large language models prompt engineering as a method for embodied cognitive linguistic representation: a case study of political metaphors in {Trump’s} discourse},
  author={Meng, Haohan and Li, Xiaoyu and Sun, Jinhua},
  journal={Frontiers in Psychology},
  volume={16},
  pages={1591408},
  year={2025},
  publisher={Frontiers Media SA}
}

@misc{ruppenhofer2016framenet,
  title={{FrameNet II}: Extended theory and practice},
  author={Ruppenhofer, Josef and Ellsworth, Michael and Petruck, Miriam RL and Johnson, Christopher R}
}

@inproceedings{an-etal-2023-coarse,
    title = "Coarse-to-Fine Dual Encoders are Better Frame Identification Learners",
    author = "An, Kaikai  and
      Zheng, Ce  and
      Gao, Bofei  and
      Zhao, Haozhe  and
      Chang, Baobao",
    editor = "Bouamor, Houda  and
      Pino, Juan  and
      Bali, Kalika",
    booktitle = "Findings of the Association for Computational Linguistics: EMNLP 2023",
    month = dec,
    year = "2023",
    address = "Singapore",
    publisher = "Association for Computational Linguistics",
    url = "https://aclanthology.org/2023.findings-emnlp.897/",
    doi = "10.18653/v1/2023.findings-emnlp.897",
    pages = "13455--13466"
}

@article{swayamdipta2017frame,
  title={Frame-semantic parsing with softmax-margin segmental {RNNs} and a syntactic scaffold},
  author={Swayamdipta, Swabha and Thomson, Sam and Dyer, Chris and Smith, Noah A},
  journal={arXiv preprint arXiv:1706.09528},
  year={2017}
}

@inproceedings{devasier-etal-2024-robust,
    title = "Robust Frame-Semantic Models with Lexical Unit Trees and Negative Samples",
    author = "Devasier, Jacob  and
      Gurjar, Yogesh  and
      Li, Chengkai",
    editor = "Ku, Lun-Wei  and
      Martins, Andre  and
      Srikumar, Vivek",
    booktitle = "Proceedings of the 62nd Annual Meeting of the Association for Computational Linguistics (Volume 1: Long Papers)",
    month = aug,
    year = "2024",
    address = "Bangkok, Thailand",
    publisher = "Association for Computational Linguistics",
    url = "https://aclanthology.org/2024.acl-long.374/",
    doi = "10.18653/v1/2024.acl-long.374",
    pages = "6930--6941"
}

@incollection{kapranov2017conceptual,
  title={Conceptual metaphors associated with climate change in corporate reports in the fossil fuels market: Two perspectives from the {United States} and {Australia}},
  author={Kapranov, Oleksandr},
  booktitle={The Role of Language in the Climate Change Debate},
  pages={90--109},
  year={2017},
  publisher={Routledge}
}

@mastersthesis{skinnemoen2009metaphors,
  title={Metaphors in climate change discourse},
  author={Skinnemoen, Jorunn},
  year={2009}
}

@book{herndl1996green,
  title={Green culture: Environmental rhetoric in contemporary America},
  author={Herndl, Carl G and Brown, Stuart C},
  year={1996}
}

@article{yang2025tale,
  title={A Tale of Identities: Environmental Identities Based on a Deliberate Metaphor Analysis of {US} Energy Companies’ Social Media},
  author={Yang, Kaiwen and Sun, Ya},
  journal={Journal of Business and Technical Communication},
  volume={39},
  number={2},
  pages={149--190},
  year={2025},
  publisher={SAGE Publications Sage CA: Los Angeles, CA}
}

@article{huang2025metaphorical,
  title={Metaphorical Framing of Climate Change in {Chinese and American} News Media: A Corpus-assisted Discourse Study},
  author={Huang, Jingyi and Liu, Ming},
  journal={Critical Arts},
  pages={1--20},
  year={2025},
  publisher={Taylor \& Francis}
}

@incollection{forgacs2022fluffy,
  title={The fluffy metaphors of climate science},
  author={Forg{\'a}cs, B{\'a}lint and Pl{\'e}h, Csaba},
  booktitle={Metaphors and analogies in sciences and humanities: Words and worlds},
  pages={447--477},
  year={2022},
  publisher={Springer}
}

@inproceedings{vallejo-etal-2025-human,
    title = "Human Interest Framing across Cultures: A Case Study on Climate Change",
    author = "Vallejo, Gisela  and
      de Kock, Christine  and
      Baldwin, Timothy  and
      Frermann, Lea",
    editor = "Rambow, Owen  and
      Wanner, Leo  and
      Apidianaki, Marianna  and
      Al-Khalifa, Hend  and
      Eugenio, Barbara Di  and
      Schockaert, Steven",
    booktitle = "Proceedings of the 31st International Conference on Computational Linguistics",
    month = jan,
    year = "2025",
    address = "Abu Dhabi, UAE",
    publisher = "Association for Computational Linguistics",
    url = "https://aclanthology.org/2025.coling-main.754/",
    pages = "11380--11398"
}

@article{lonneker2008hamburg,
  title={The hamburg metaphor database project: issues in resource creation},
  author={L{\"o}nneker-Rodman, Birte},
  journal={Language resources and evaluation},
  volume={42},
  number={3},
  pages={293--318},
  year={2008},
  publisher={Springer}
}

@inproceedings{wachowiak-etal-2022-drum,
    title = "Drum Up {SUPPORT}: Systematic Analysis of Image-Schematic Conceptual Metaphors",
    author = "Wachowiak, Lennart  and
      Gromann, Dagmar  and
      Xu, Chao",
    editor = "Ghosh, Debanjan  and
      Beigman Klebanov, Beata  and
      Muresan, Smaranda  and
      Feldman, Anna  and
      Poria, Soujanya  and
      Chakrabarty, Tuhin",
    booktitle = "Proceedings of the 3rd Workshop on Figurative Language Processing (FLP)",
    month = dec,
    year = "2022",
    address = "Abu Dhabi, United Arab Emirates (Hybrid)",
    publisher = "Association for Computational Linguistics",
    url = "https://aclanthology.org/2022.flp-1.7/",
    doi = "10.18653/v1/2022.flp-1.7",
    pages = "44--53"
}

@article{ahrens2022referential,
  title={Referential and evaluative strategies of conceptual metaphor use in government discourse},
  author={Ahrens, Kathleen and Zeng, Winnie Huiheng},
  journal={Journal of Pragmatics},
  volume={188},
  pages={83--96},
  year={2022},
  publisher={Elsevier}
}

@inproceedings{wachowiak-gromann-2023-gpt,
    title = "Does {GPT}-3 Grasp Metaphors? {I}dentifying Metaphor Mappings with Generative Language Models",
    author = "Wachowiak, Lennart  and
      Gromann, Dagmar",
    editor = "Rogers, Anna  and
      Boyd-Graber, Jordan  and
      Okazaki, Naoaki",
    booktitle = "Proceedings of the 61st Annual Meeting of the Association for Computational Linguistics (Volume 1: Long Papers)",
    month = jul,
    year = "2023",
    address = "Toronto, Canada",
    publisher = "Association for Computational Linguistics",
    url = "https://aclanthology.org/2023.acl-long.58/",
    doi = "10.18653/v1/2023.acl-long.58",
    pages = "1018--1032"
}

@article{mason2004cormet,
  title={{CorMet}: A computational, corpus-based conventional metaphor extraction system},
  author={Mason, Zachary J},
  journal={Computational linguistics},
  volume={30},
  number={1},
  pages={23--44},
  year={2004}
}

\appendix
\section{Appendix}

\subsection{Pre-training models finetuning details}

\subsubsection{Semantic frame prediction}
\label{app:sem_frame_prediction}

\textbf{Fine-tuning.} We use input length of 256, batch size of 32, Adam W optimization, and 2 warm-up epochs.  All models are trained for the maximum of 10 epochs with patience of 2 as determined by absence of improvement in terms of binary-F1. The starting learning rate is $2e^{-5}$ with linear learning rate scheduler and weight decay of 0.1.

\subsubsection{Source domain prediction}
\label{app:source_domain_prediction}

\textbf{Data.} We use the large version (ENGLISH large) of the LCC Metaphor Dataset (large version) \citep{mohler2016introducing}. To ensure high quality of training data, we only use human-validated subsets of the dataset (ANN, REC, SYS), and only the samples where at least half of the annotators regarding the source domain, or there is only one annotation. We also remove those source domains which have less than 3 samples, to ensure valid evaluation and testing. This results in 16722 samples, out of which 11154 have single annotations, and 5568 are annotated by at least two people (with a high agreement in terms of Krippendorf's $\alpha$ 0.76 and Fleiss $\kappa$ of 0.76).

\textbf{Fine-tuning details.} All models have the same hyperparameters in terms of input length (256), batch size of 32, Adam W optimization, and 2 warm-up epochs.  All models are trained for the maximum of 30 epochs with patience of 5 as determined by absence of improvement in terms of macro-F1. The starting learning rate is $2e^{-5}$ with linear learning rate scheduler and weight decay of 0.1.

\subsubsection{Metaphor prediction}
\label{app:metaphor_prediction}

\textbf{Data.}
\noindent\textbf{VUA-18}: we use the original splits of 6323 (training)/1550 (development)/2694 (test). We use all samples (metaphors across all parts-of-speech), the average metaphor rate being 22\%. We use the training split for fine-tuning, and report the results on the test set.

\noindent\textbf{TroFi}: we use the whole corpus (3737 sentences) for evaluation (no fine-tuning, in zero-shot way). The corpus focuses on verbs and has a metaphor rate of around 43\%.

\noindent\textbf{MOH-X}: we use the whole corpus (647 sentences) for evaluation for evaluation (no fine-tuning, in zero-shot way). The corpus focuses on verbs and has a metaphor rate of around 49\%.

\textbf{Fine-tuning details.} For fair comparison with Melbert, we reproduce fine-tuning parameters used in the original implementation \citep{choi2021melbert}. In particular, we use the learning rate of 3e-05 with 2 warm-up epochs and the drop ratio of 0.2, and train for 3 epochs. 

\subsection{Classifier prompts}

\subsubsection{Semantic frame classification}
\label{app:sem_frame_prompt}

We use the following prompt, where \textbf{labels} is substituted with the list of semantic frames and their definitions:

\begin{prompt}

You are a linguist specializing in semantic frames (FrameNet).\\

Please choose the semantic frame out of the following list: \textbf{labels}. Do not use any other labels, and do not change the wording of the label. Do not remove "\_" in the label if it exists.\\

Please return a json object which consists of the following field:\\

"frame": one of the values from the list.\\

Do not output anything else. Do not output any reasoning steps or explanations.\\
\end{prompt}

\subsubsection{Source domain classification}
\label{app:source_domain_prompt}

We use the following prompt, which is based on human annotation instructions (see \Cref{app:annotation_task}), where \textbf{labels} is substituted with the list of source domains:

\begin{prompt}
    You are a linguist specializing in metaphors. You will be given a metaphor and a sentence it occurs in. You will be asked to identify the source domain of the metaphor.\\

    A metaphor is when you describe something by saying it is something else, even though it is not literally true. For example, we can say "They are forced to make a decision" while there is no actual physical force applied to them.\\

    When using a metaphor, we are carrying over associations from a more tangible and specific source domain (such as physical force or pressure) to a more abstract domain (such as obligation).\\

    In this study, you will need to identify the source domain of the metaphor. It is helpful to remember that the source domain is usually a more specific, physical thing (force, pressure) while the target domain is more abstract (obligation).\\

    Please choose the source domain out of the following list: \textbf{labels}. Do not use any other labels, and do not change the wording of the label, Do not remove "\_" in the label if it exists.\\

    Please return a json object which consists of the following field:\\

    "source": one of the values from the list.\\

    Do not output anything else. Do not output any reasoning steps or explanations.\\
\end{prompt}

\subsubsection{Metaphor prediction}
\label{app:met_pred_prompt}

We use the MIP protocol for the prompt, which our preliminary experiments have shown to be more effective than other metaphor identification protocols such as SPV or CMT.

\begin{prompt}

You are a linguist specializing in metaphors. You will be given a sentence and asked to find all metaphors in it.\\
Please extract all text spans that have a metaphorical meaning. \\

Go through the following steps to determine if a word is used in a metaphorical meaning:\\

1. Read the entire sentence to establish a general understanding of the meaning.\\
2. (a) Establish the word's meaning in context, that is,
how it applies to an entity, relation, or attribute in the situation evoked
by the text (contextual meaning). Take into account what comes before
and after the lexical unit.\\
(b) Determine if the target word has a more basic contemporary
meaning in other contexts than the one in the given context. For our
purposes, basic meanings tend to be\\
—More concrete; what they evoke is easier to imagine, see, hear, feel,
smell, and taste.\\
—Related to bodily action.\\
—More precise (as opposed to vague)\\
—Historically older.\\
Basic meanings are not necessarily the most frequent meanings of the
target word.\\
(c) If the target word has a more basic current–contemporary meaning in
other contexts than the given context, decide whether the contextual
meaning contrasts with the basic meaning but can be understood in
comparison with it.\\
3. If yes, the word is metaphorical. Otherwise it is literal.\\

Please return a json object which consists of the following field:\\

metaphors: a list of extracted metaphor spans.\\

Do not output any explanations or reasoning steps.\\

\end{prompt}

To evaluate the extracted spans, we check if they overlap with golden spans in the dataset. We consider partial overlaps to be correct.

\subsection{Most often confused semantic frames}
\label{app:confusion_sem_frames}

The table below shows the pairs of most often confused semantic frames by RoBERTA SEQ, as well as their raw counts in the test set (out of 6714 samples).

\begin{table}[h]
\centering
\small
\begin{tabular}{llc}
\hline
\textbf{Sem. frame 1} & \textbf{Sem. frame 2} & \textbf{Count} \\
\hline
Buildings & Locale\_by\_use & 23 \\
Locating & Becoming\_aware & 14 \\
Possibility & Capability & 13 \\
Calendric\_unit & Measure\_duration & 12 \\
Ride\_vehicle & Bringing & 11 \\
Stage\_of\_progress & Temporal\_collocation & 10 \\
Posture & Change\_posture & 10 \\
Have\_associated & Possession & 9 \\
Capability & Possibility & 8 \\
Touring & Travel & 8 \\
\hline
\end{tabular}
\caption{Frame pairs and counts}
\label{tab:frame_pairs}
\end{table}

\subsection{Full results for metaphor classifier performance}

\begin{table*}[th]
    \centering
    \small
    \setlength{\tabcolsep}{5pt} 
    \begin{tabular}{lcccc|cccc|cccc}
    
        \toprule
        & \multicolumn{4}{c}{\textbf{VUA-18}} & \multicolumn{4}{c}{\textbf{TroFi}} & \multicolumn{4}{c}{\textbf{MOH-X}} \\
        \cmidrule(lr){2-5} \cmidrule(lr){6-9} \cmidrule(lr){10-13}
        & \textbf{Acc} & \textbf{P} & \textbf{R} & \textbf{F1} & \textbf{Acc} & \textbf{P} & \textbf{R} & \textbf{F1} & \textbf{Acc} & \textbf{P} & \textbf{R} & \textbf{F1}\\
        \midrule
        Random baseline & 0.499 & 0.142 & 0.501 & 0.222 & 0.500 & 0.436 & 0.501 & 0.466 & 0.484 & 0.468 & 0.505 & 0.486 \\
        \midrule
        MelBert & 0.938 & 0.778 & 0.786 & 0.782 & 0.593 & 0.521 & 0.801 & 0.631 & 0.800 & 0.753 & 0.871 & 0.806 \\
        FrameBert & 0.933 & 0.757 & 0.776 & 0.766 & 0.595 & 0.524 & 0.761 & 0.620 & 0.769 & 0.720 & 0.859 & 0.780 \\
        ConceptFrameMet & 0.931 & 0.732 & \textbf{0.806} & 0.767 & \textbf{0.605} & \textbf{0.532} & 0.787 & \textbf{0.634} & \textbf{0.803} &  0.746 & \textbf{0.899} & \textbf{0.814} \\
        \midrule
        Gemini 2.5 & 0.786 & 0.304 & 0.390 & 0.341 & 0.601 & 0.528 & 0.789 & 0.633 & 0.847 & 0.809 & 0.893 & 0.849 \\
        Claude Sonnet 4.0 & 0.827& 0.367 & 0.301 & 0.330 & 0.602 & 0.526 & 0.852 & 0.651 & 0.806 & 0.739 & 0.926 & 0.822 \\
        \bottomrule
    \end{tabular}
    \caption{Full results for metaphor prediction task for pre-traimed models fine-tuned on VUA-18 dataset. The reported metrics are binary (for the metaphor class).}
    \label{tab:full_metaphor_results}
\end{table*}

\subsection{Source prediction performance depending on the class frequency}

\begin{figure}
    \centering
    \includegraphics[width=1\linewidth]{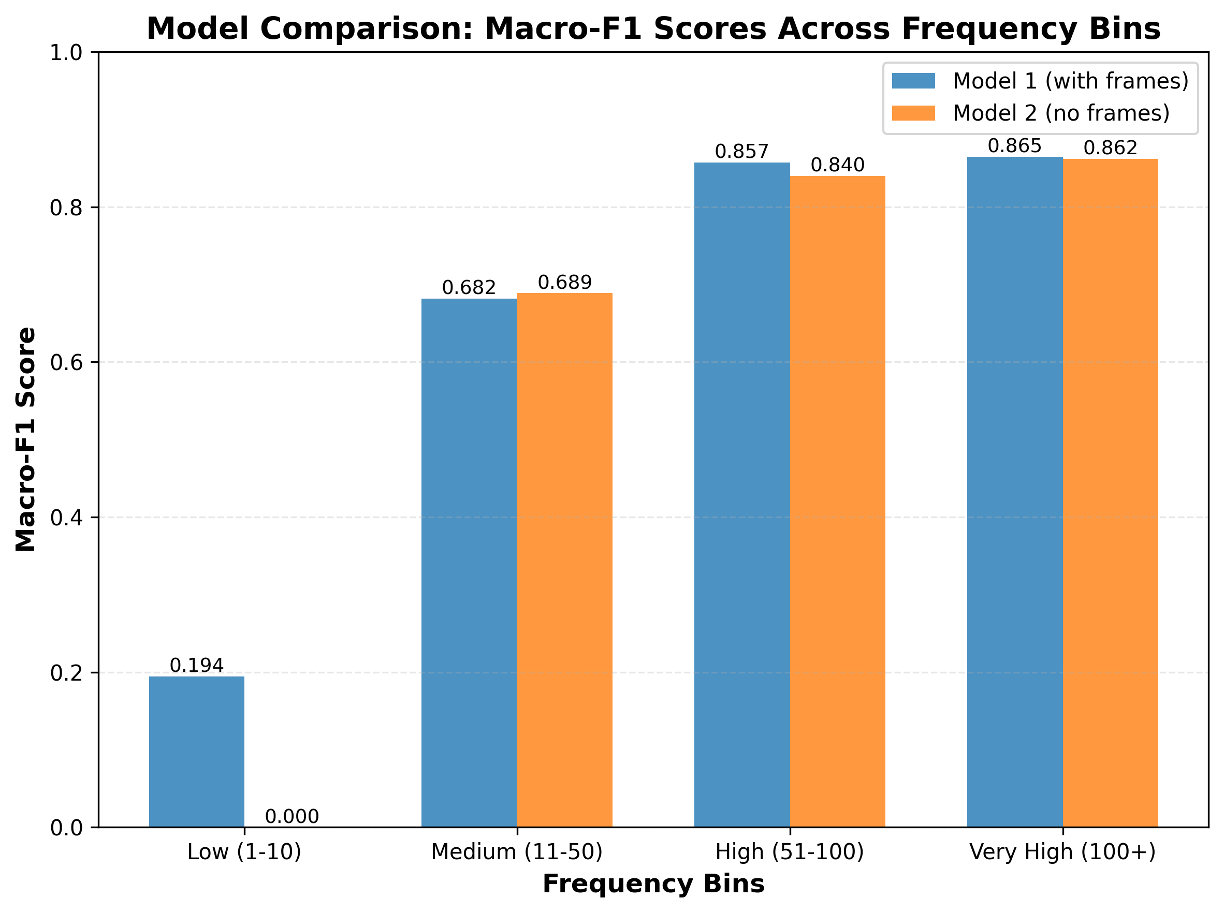}
    \caption{Performance comparison on test set for classes with different frequency in the training set.}
    \label{fig:binned_results}
\end{figure}

\Cref{fig:binned_results} compared performance of SEP-style RoBERTa models with and without semantic frames on test set for classes with different frequency in the training set.

\subsection{Collecting the climate change corpus and generic corpus}
\label{app:term_distrubution}

We use a corpus of New York Times articles published between 1986 and 2020 and extract all paragraphs which mention ``climate change'' or ``global warming''. Next, we filter them using a ClimateBERT pre-trained model \citep{webersinke2021climatebert} which was fine-tuned to detect climate-related texts \citep{bingler2023cheaptalk}, and, additionally,  use the prompt from \citet{vallejo-etal-2025-human} with Gemini 2.5 Flash to detect if a paragraph focuses on climate change (rather than just contains the keywords). 

\begin{figure}[htbp]
    \centering
    \includegraphics[width=0.45\textwidth]{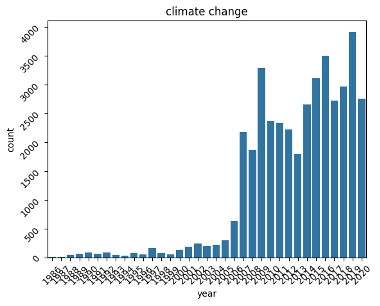}
    \hfill
    \includegraphics[width=0.45\textwidth]{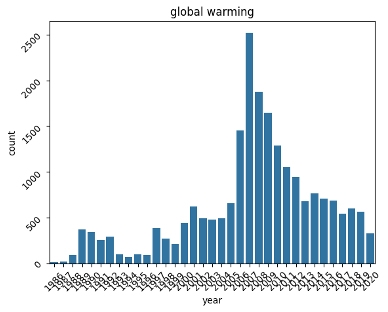}
    \caption{Distribution of mentions of ``climate change'' or ``global warming'' across 1986-2020}
    \label{fig:term_distribution}
\end{figure}

\Cref{fig:term_distribution} shows distribution of mentions of ``climate change'' or ``global warming''  in New York Times over the period from 1986 to 2020. 

For saliency analysis, we also collect a generic corpus randomly sampled from the same NYT data set ensuring that the paragraphs do not contain our climate keywords. We sample the same number of paragraphs per year as in the climate corpus (47K in total).

\subsection{Top source domains in climate corpus}
\label{app:top_source_domains}

\Cref{tab:top_domains_refs} shows the most salient source domains in NYT climate corpus, together with their examples, alternative names, and selected theoretical studies that identified them.

\begin{table*}[t]
    \centering
    \small
    \setlength{\tabcolsep}{4pt} 
   
    \label{tab:placeholder_label}
    \begin{tabular}{c | c | c | c }
        \hline
        Source domain & Alternative names & Examples & Reference \\
        \midrule
        STRUGGLE  & CHALLENGE & fight, confront & \citet{kapranov2017conceptual}\\
        WAR &  BATTLE & battle, war, on the front lines  & \citet{skinnemoen2009metaphors} \\
        OBJECT HANDLING  & OBJECT & push (for), press, handle, take (action) & \citet{herndl1996green} \\
        HUMAN BODY & BODY & face, confront, embrace, at the heart of  & \citet{nerlich2012metaphors} \\
        PATHWAY & PATH & passing, path  & \citet{yang2025tale} \\
        BUILDING & CONSTRUCTION & build, lay a foundation  & \citet{huang2025metaphorical} \\
        FURNISHINGS & & climategate, take a back seat  & \citet{forgacs2022fluffy} \\
        VISION & & view, presage  &  \\
        FORWARD MOVEMENT & & ahead, progress, gain  & \citet{skinnemoen2009metaphors} \\
        BACKWARD MOVEMENT &  & backwards, rollback  & \citet{skinnemoen2009metaphors} \\

    \end{tabular}
     \caption{Top 10 salience source domains in NYT climate corpus, together with their alternative names in the literature, examples and references. For brevity we provide only one reference per source domain.}
     \label{tab:top_domains_refs}
\end{table*}

\subsection{Immigration source domains}
\label{app:immi_source_domains}

The corpus provided by \citet{mendelsohn-budak-2025-people} contains tweets automatically annotated with predicted metaphoricity scores across seven dehumanization source domains. The authors also release a smaller dataset with 200 tweets per source domain, where each tweet is judged by 10 human annotators as containing or not containing a metaphor from that particular domain.

As the larger dataset only contains automatically predicted probabilities, we convert them to (potential) source domain labels as follows.  We regard a tweet $t$ as potentially containing a metaphor with a particular source domain $s$ if the predicted probability of that source domain $P(s|t)$ is larger than the average metaphoricity score $\bar{m}_s = \frac{1}{|T_s|} \sum_{t' \in T_s} m_{t'}$ of tweets that were judged by annotators as belonging to that source domain (these cut-off scores are in \Cref{app:immi_source_domains}).

For precision, we only use tweets which have the same source domain annotation at both tweet (original predictions) and metaphor level (our model). Finally, since a metaphor with a WATER, ANIMAL or WAR source can refer to another, irrelevant target rather than immigration (for example, as in \textit{left-\textbf{winged} politicians}), we further filter out tweets where the metaphor does not refer to immigrants. 

To filter out metaphors which have the source domain we are interested in but the wrong (irrelevant) target, we first use Claude Sonnet 4.5 with the prompt below, where we ask it to resolve the target which the specific metaphor refers to:

\begin{prompt}

You are a linguist specializing in metaphors. You will be given a sentence and a metaphor in it. Identify and out out the target of the metaphor, i.e. the entity, person, or object it refers to. For example, given the following input: \\
\\
"text": "Immigrants are flooding into our country and ruining our economy" \\
"metaphor": "flooding" \\
\\
output the following: \\
"target": "immigrants"\\
\\
Make sure to use json format with this field. Do not output anything else, no explanations! You are NOT allowed to say "no metaphor is present" since the metaphor is given to you!

\end{prompt}

In this way, we collect a list of target domains used in tweets, such as ``undocumented immigrant'', ``illegals'', ``Liberals'', ``economy'', or ``Trump'', and filter it out manually to collect a list of expressions that are likely to refer to the target domain we are interested in (immigration). In particular, we arrive at the following list of substrings which then match against the predicted target domains: ``immi'', ``illegal'', ``alien'', ``ICE'', ``migr'', ``foreign'', ``refug'', ``detention'', ``border'', ``asylum'', ``famil''.

\Cref{tab:immi_source_domains} below shows the source domains used in metaphorical framing of tweets about immigration, their definitions, the average metaphoricity scores from the annotated corpus by \citep{mendelsohn-budak-2025-people}, the number of tweets from the unlabeled part of that corpus with metaphoricity scores about that cut off, as well as the final number of tweets used for analysis after filtering.

\begin{table*}[ht]
    \centering
    \begin{tabular}{l|cccc}
         Source domain & Examples & Avg metaphoricity score & \# of candidates & \# after filtering \\
         \midrule
         ANIMAL &  shelter, cage, swarm & 0.3045 & 9845 & 544 \\
         WAR & army, hordes & 0.3974 & 16455 &  3695 \\
         WATER & flood, tide, pour &  0.4375 & 3661 & 1402 \\
    \end{tabular}
    \caption{Statistics for ANIMAL, WAR and WATER source domains}
    \label{tab:immi_source_domains}
\end{table*}

\subsection{Annotation task details}
\label{app:annotation_task}

\subsubsection{Task overview}
We collected human annotations for metaphor source domains using a custom web-based annotation platform.\footnote{Available at \url{https://metaphor-annotation-source.onrender.com/}} Annotators were presented with sentences containing highlighted metaphors and asked to identify up to 3 source domains from a provided list of options for each metaphor. The annotation interface included a mandatory 10 second reading period before options appeared, and copy-paste was not allowed, designed to encourage careful consideration of the metaphorical usage in context and to avoid AI-powered tools. The order of samples within the batch was randomized to avoid batch effects. 

\subsubsection{Annotator recruitment}
Annotators were recruited through Prolific (\url{https://www.prolific.com/}), a crowdsourcing platform commonly used for academic research. The study collected participants' Prolific IDs, study IDs, and session IDs to enable proper tracking and compensation. All participants were required to provide informed consent through their Prolific ID before proceeding to the annotation task. However, all participants data (Prolific IDs) was anonymised after collection and quality checks. Participants were required to be native English speakers, with English listed as their first language. Eligible annotators reported their country of birth as either the United Kingdom or the United States. To ensure a high level of reading comprehension and annotation quality, participants were additionally required to have completed at least an undergraduate degree (BA/BSc or equivalent), with many holding graduate or doctoral degrees. Only Prolific users with an approval rate between 98\% and 100\% were permitted to participate.

This study was approved by the Human Ethics Committee (Reference No.\ 2025-32051-65749-4) and has been conducted according to the corresponding ethical standards.

\subsubsection{Annotation instructions}
\begin{figure*}[ht]
\centering
\includegraphics[width=0.95\textwidth]{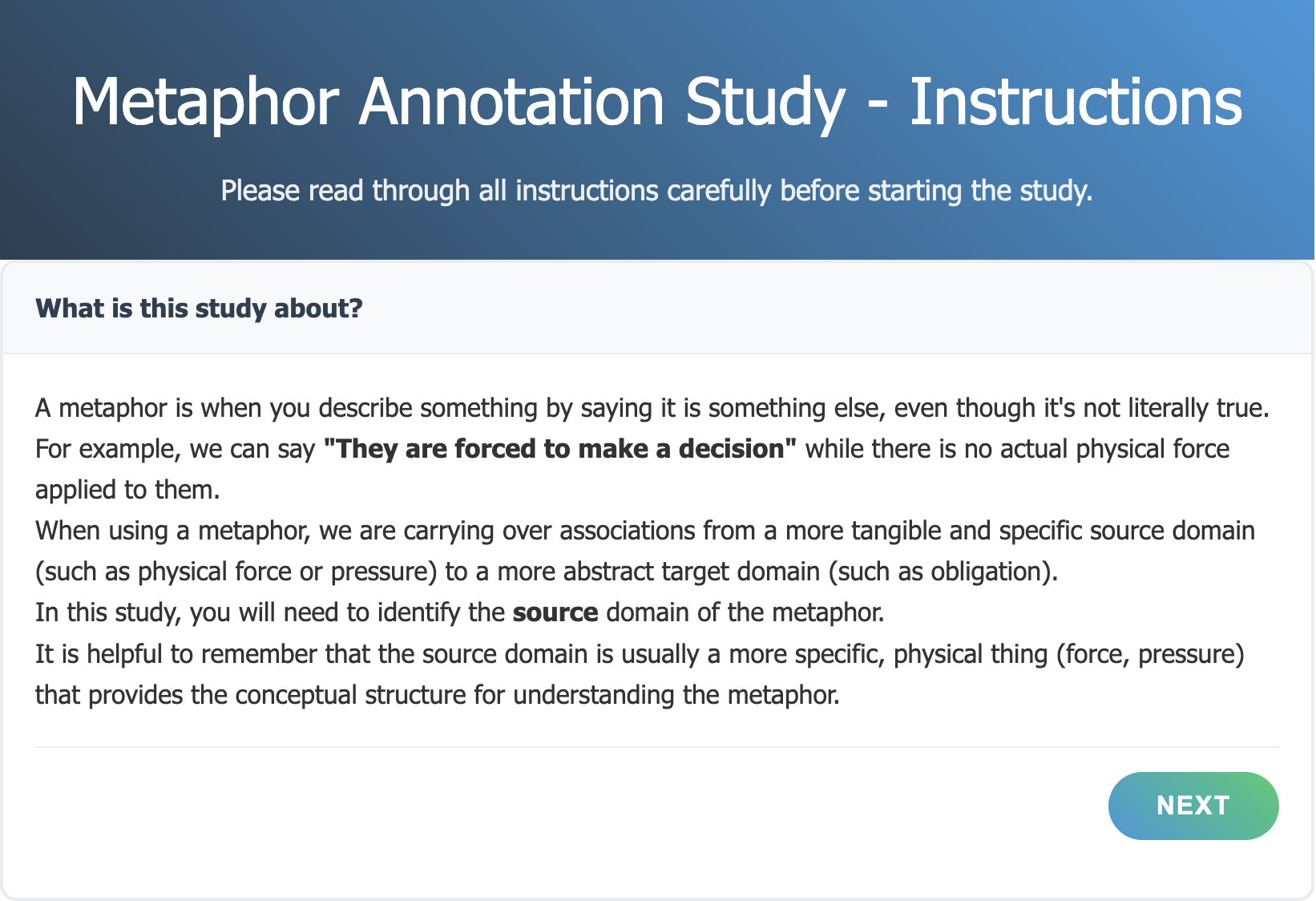}

\vspace{10pt}

\includegraphics[width=0.95\textwidth]{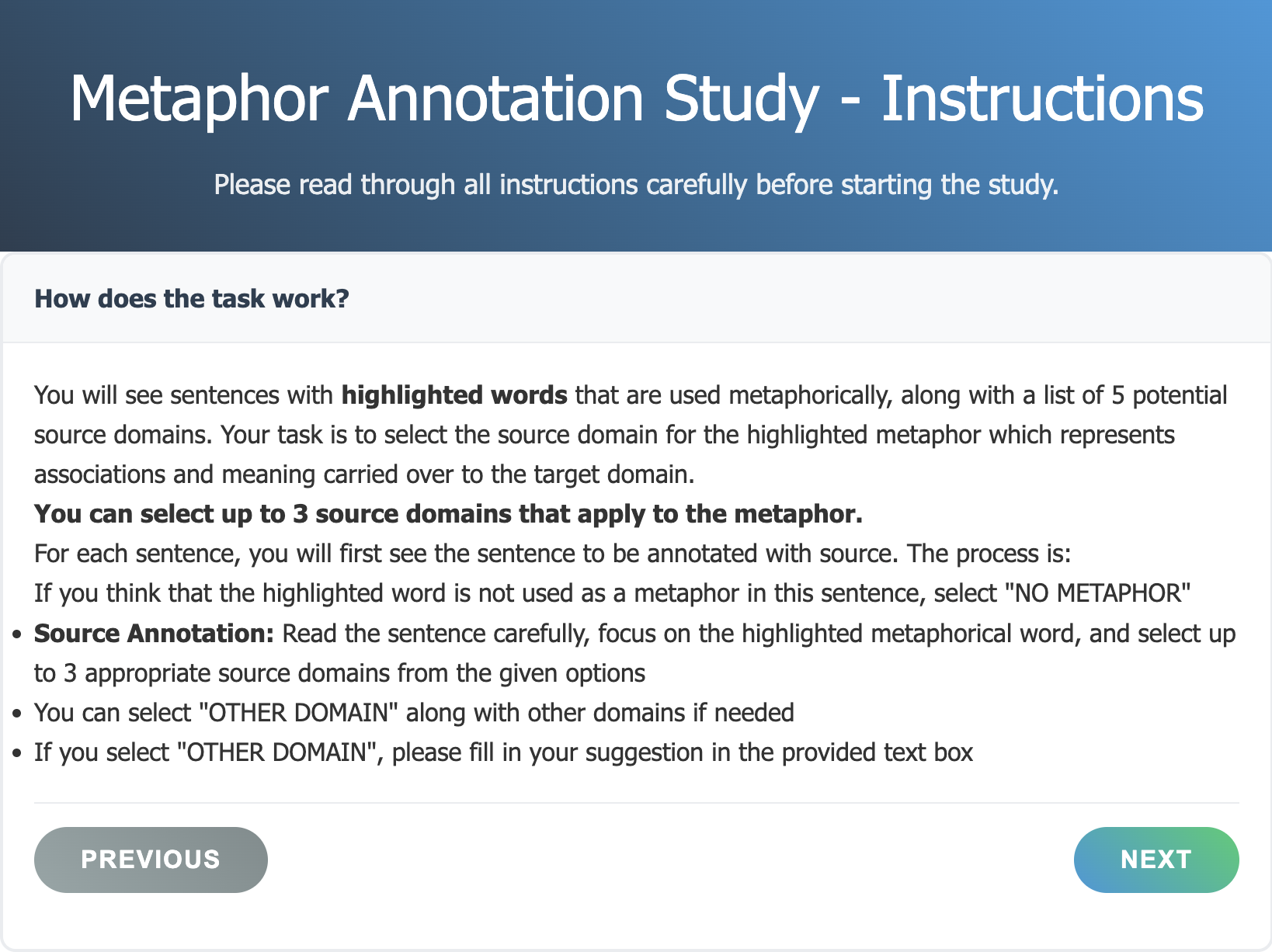}

\caption{Screenshots of the annotation instruction interface. Annotators were guided through an interactive tutorial explaining (top) the concept of metaphor source domains and (bottom) the annotation task workflow before beginning comprehension checks.}
\label{fig:annotation_instructions}
\end{figure*}

\Cref{fig:annotation_instructions} features the instructions provided to annotators during the interactive tutorial phase.

\subsubsection{Quality assurance}

We implemented a multi-stage quality assurance protocol to ensure high annotation reliability. Prior to beginning the main annotation task, participants were shown five example annotations and then completed five comprehension (attention check) questions with known correct answers. To pass, participants were required to answer at least three out of five questions correctly (i.e., a maximum of two errors). Participants who failed the comprehension check were permitted one retry after reviewing the task instructions. A second failure resulted in permanent rejection and participants were instructed to return their submission on Prolific.

During the annotation task, participants were required to complete each sample sequentially before proceeding to the next. For each sample, annotators could select a maximum of three source domains. They could also select \textit{No Metaphor}, which was mutually exclusive with all other options, or \textit{Other Domain}, which prompted them to manually specify a domain not listed among the predefined options.

The annotation platform explicitly tracked three rejection scenarios:
\begin{enumerate}
    \item \textbf{First comprehension check failure}: participants were allowed to retry after reviewing the instructions.
    \item \textbf{Second comprehension check failure}: participants were permanently rejected and instructed to return their submission on Prolific.
    \item \textbf{Incomplete submissions}: participants who did not complete all 50 annotations in their assigned batch were automatically instructed to return their submission.
\end{enumerate}

Participants were informed during the consent process that failing comprehension checks would result in submission return or rejection, and that repeated failures would lead to ineligibility for future tasks within the project unless manually re-approved.

\subsubsection{Compensation}

Each annotation batch was completed by four independent annotators. Participants were allotted up to 45 minutes to complete a batch of up to 50 annotations. Compensation was set at an hourly rate of \pounds12, in line with Prolific’s recommended fair pay guidelines and well above the local standard minimum wage.

\subsubsection{Annotation agreement and adjudication}

\Cref{tab:iaa} shows averaged pair-wise annotation agreement between 4 annotators (6 comparisons) within each batch, as well as overall (mean) agreement across batches. Since annotators were allowed to choose multiple labels per sample, agreement is defined as at least one of those labels being the same, i.e. any overlap between the selected labels. To evaluate how consistent the labels were among all four annotators, we also report the rate of strong majority vote agreement, i.e. percentage of samples where the same (and only one) label was chosen by at least three out of four annotators.

\begin{table}
    \centering
    \begin{tabular}{c|c|c}
         Batch & Agreement rate & Majority vote rate \\
         \midrule
         1 & 0.68 & 0.74 \\
        2 & 0.66 & 0.73 \\
        3 & 0.71 & 0.70 \\
        4 & 0.75 & 0.72 \\
        5 & 0.73 & 0.75 \\
        6 & 0.55 & 0.68 \\
        7 & 0.77 & 0.82\\
        8 & 0.61 & 0.73 \\
        9 & 0.69 & 0.63 \\
        \midrule
        Mean & 0.68 & 0.72\\
    \end{tabular}
    \caption{Per batch IAA rate}
    \label{tab:iaa}
\end{table}

Whenever the majority vote could not be established (or the majority vote was OTHER DOMAIN, i.e. not one of the suggested classes), the labels were adjudicated by the first author of the paper, a postdoctorate researcher with background in linguistics and pragmatics. The adjudicator considered the selected labels, as well as the labels for samples containing related words (for example, if the label for ``vision'' could not be established, the labels for ``visions'' were checked). In cases where none of the five suggested labels was appropriate, the adjudicator was allowed to choose the label from the full list of source domains. The adjudicator also ensured that the source domains were assigned consistently between multiple samples containing the same metaphor.

\subsection{Most often confused source domains}
\label{app:confusion}

To get the list of the most often confused labels, we calculate normalized point-wise mutual information and weight the score by the frequency of label pairs:

\[
\text{NPMI}_w(a,b) = \frac{\log \frac{p(a,b)}{p(a)p(b)}}{-\log p(a,b)} \times \log c(a,b)
\]

where c(a,b) is the co-occurrence count of labels \textit{a} and \textit{b}.

\begin{table*}[h]
\centering
\begin{tabular}{llrrr}
\toprule
\textbf{Label A} & \textbf{Label B} & \textbf{Count} & \textbf{NPMI} & \textbf{NPMI\textsubscript{w}} \\
\midrule
BATTLE & WAR & 33 & 0.751 & 2.627 \\
COMPETITION & STRUGGLE & 30 & 0.554 & 1.884 \\
OBJECT HANDLING & PHYSICAL OBJECT & 37 & 0.423 & 1.528 \\
BATTLE & STRUGGLE & 17 & 0.393 & 1.114 \\
STRUGGLE & WAR & 16 & 0.367 & 1.019 \\
MOVEMENT & VEHICLE & 19 & 0.286 & 0.843 \\
BARRIER & PATHWAY & 9 & 0.374 & 0.822 \\
FORWARD MOVEMENT & UPWARD MOVEMENT & 5 & 0.390 & 0.628 \\
PHYSICAL HARM & TOOL & 4 & 0.433 & 0.601 \\
LIGHT & VISION & 9 & 0.264 & 0.579 \\
\bottomrule
\end{tabular}
\caption{Top co-occurring label pairs by weighted NPMI (within the labels chosen by the same annotator)}
\label{tab:npmi_within}
\end{table*}

\begin{table*}[h]
\centering
\begin{tabular}{llrrr}
\toprule
\textbf{Label A} & \textbf{Label B} & \textbf{Count} & \textbf{NPMI} & \textbf{NPMI\textsubscript{w}} \\
\midrule
COMPETITION & STRUGGLE & 17 & 0.658 & 1.863 \\
BATTLE & WAR & 12 & 0.716 & 1.778 \\
OBJECT HANDLING & PHYSICAL OBJECT & 21 & 0.442 & 1.344 \\
BARRIER & PATHWAY & 9 & 0.494 & 1.086 \\
LIGHT & VISION & 11 & 0.443 & 1.062 \\
BUILDING & SHAPE & 7 & 0.429 & 0.835 \\
STRUGGLE & WAR & 8 & 0.388 & 0.807 \\
BATTLE & STRUGGLE & 7 & 0.367 & 0.715 \\
FORCEFUL EXTRACTION & OBJECT HANDLING & 7 & 0.357 & 0.695 \\
AVERSION & STRUGGLE & 4 & 0.450 & 0.624 \\
\bottomrule
\end{tabular}
\caption{Top co-occurring label pairs by weighted NPMI (within the labels chosen by all annotators)}
\label{tab:npmi_between}
\end{table*}

\subsection{Prompts for end-to-end detection of salient semantic frames within WATER source domain}
\label{app:end_to_end_prompt}

\subsubsection{System prompt}

\begin{prompt}
You are a linguist specializing in discourse analysis. You will be given two collection of tweets - by liberals and by conservatives. You need to analyse how WATER metaphors are used by these sides. Output the most salient (prominent) semantic frames used for WATER metaphors by liberal and conservative sides.\\

Please choose the semantic frame out of the following list: \textbf{\{labels\}}. Do not use any other labels, and do not change the wording of the label. Do not remove "\_" in the label if it exists.\\

Please return two lists, each containing salient semantic frames and their examples. COPY specific examples from tweets.
'''
\end{prompt}

\subsubsection{User prompt}

\begin{prompt}
What are the prevalent semantic frames used in WATER metaphors by liberals and conservatives?\\

Liberal tweets: \textbf{\{liberals\}}\\

Conservative tweets: \textbf{\{conservatives\}}\\
\end{prompt}

\subsection{Examples of end-to-end LLM metaphor analysis}

\subsubsection{ANIMAL source domain}
\label{app:end_to_end_animal}

\begin{prompt}
    
**Liberal Semantic Frames:**

1. **Abusing** - Liberals frequently use animal metaphors to describe how immigrants are being mistreated, with phrases like "treating immigrants like animals," "caging children like animals," and "treating them worse than dogs"

2. **Containing** - Extensive use of metaphors describing detention facilities as "cages," "kennels," "dog pounds," and "pens" where immigrants are held

3. **Hunting** - Metaphors describing ICE and border patrol activities as "hunting down immigrants," "stalking," and "preying on" vulnerable populations

4. **Dehumanization** - References to how Trump and officials call immigrants "animals," "snakes," "cockroaches," and other dehumanizing terms

5. **Protecting** - Calls to protect immigrants from being treated like animals, with references to animal welfare standards being better than immigrant treatment

**Conservative Semantic Frames:**

1. **Categorization** - Frequent use of "animals" to categorize MS-13 gang members and violent criminals, distinguishing them from law-abiding citizens

2. **Being\_in\_category** - Classifying certain immigrants as "illegal aliens" versus legal immigrants, often using animal terms for the former group

3. **Swarming** - Describing immigration as animal-like mass movement with terms like "swarms," "herds," "flocks," and "stampedes" of immigrants

4. **Hunting** - References to "hunting down" and "rounding up" illegal immigrants for deportation

5. **Proliferating\_in\_number** - Using breeding metaphors to describe immigrant population growth, with terms like "breeding like rabbits" and concerns about demographic changes

\subsubsection{WATER source domain - based on tweets that contain WATER metaphors}

\label{app:end_to_end_water1}

**Liberal Side - Most Salient Semantic Frames:**

1. **Fluidic\_motion** - This is the most prominent frame, with liberals using water metaphors to describe the movement and flow of immigrants in neutral or sympathetic terms. Examples include "flow of migrants," "stream of immigrants," "waves of immigration," and "tide of migrants." These metaphors often emphasize the natural, ongoing nature of migration.

2. **Motion** - Closely related to fluidic motion, this frame captures the directional movement of people, often described as "pouring in," "flooding," or "streaming" across borders, but typically used to critique anti-immigration policies rather than the immigrants themselves.

3. **Catastrophe** - Used to describe the humanitarian crisis and dangerous conditions migrants face, such as "drowning" in the Rio Grande, being "swept away," or facing "floods" of persecution in their home countries.

**Conservative Side - Most Salient Semantic Frames:**

1. **Fluidic\_motion** - Also the most prominent frame for conservatives, but used with negative connotations to describe an overwhelming, uncontrolled influx. Examples include "flood of illegals," "pouring across the border," "stream of illegal aliens," and "tide of illegal immigration." The emphasis is on the volume and lack of control.

2. **Catastrophe** - Used to describe the perceived negative impact of immigration on American society, such as the country being "flooded," "swamped," or "drowning" in illegal immigrants, suggesting an overwhelming disaster.

3. **Emptying** - This frame appears in contexts of "draining" resources, the economy, or social services due to immigration, suggesting that immigrants are depleting American resources like water draining from a container.

\subsubsection{WATER source domain - based on tweets that do NOT contain WATER metaphors}
\label{app:end_to_end_water2}

**Liberal Tweets - Most Salient Water Metaphor Semantic Frames:**

1. **Fluidic\_motion**
   - Examples: "immigrants...pour in and infest our country", "flooding our country with immigrants", "immigrants are swarming to the border", "migrants fleeing violence in El Salvador still plan for the U.S."

2. **Catastrophe** 
   - Examples: "more hopeless. I feel like we are about to get buried by a tsunami...Iran/Roe/Immigration", "We're not being invaded from any direction, we're not under attack by immigrants"

3. **Abounding\_with**
   - Examples: "Trump refers to immigrants as an infestation 46 times", "immigrants...pour in and infest our country", "calling migrants invaders and an infestation"

**Conservative Tweets - Most Salient Water Metaphor Semantic Frames:**

1. **Fluidic\_motion**
   - Examples: "illegals are flooding states", "Democrats divert 137B tax\$ ANNUALLY, to support \& defend illegals", "illegals flooding the country", "flood our country with immigrants", "illegals pouring in"

2. **Catastrophe**
   - Examples: "NATIONAL EMERGENCY: 2,000+ Migrants Quarantined After Bringing Disease", "This is an invasion", "illegals flooding your border overwhelming your system"

3. **Scarcity**
   - Examples: "illegals cost America over 300 Billion", "illegals are costing us billions", "Each illegal immigrant costs taxpayers \$70k per year", "illegals cost US well over 300 Billion a yr"

\end{prompt}

\end{document}